%% file: root.tex
\documentclass[letterpaper, 10 pt, conference]{ieeeconf}  %

\IEEEoverridecommandlockouts                              %
\let\labelindent\relax

\usepackage{booktabs}       %
\usepackage{xspace}
\usepackage{wrapfig}
\usepackage{array}
\usepackage[font={small}]{caption}
\usepackage{enumitem}
\usepackage{multirow}
\usepackage{amssymb}
\usepackage{longtable}
\usepackage{threeparttable}
\usepackage[T1]{fontenc}
\usepackage{footnote}
\makesavenoteenv{tabular}
\makesavenoteenv{table}

\newcommand\mybar{\kern1pt\rule[-\dp\strutbox]{.8pt}{\baselineskip}\kern1pt}

\usepackage{hyperref}
\hypersetup{
    colorlinks=true,
    linkcolor=black,
    citecolor=black,
    filecolor=cyan,
    urlcolor=blue
}

\usepackage{misc/onimage}
\tikzset{
    image label/.style={
        every node/.style={
            fill=black,
            text=white,
            font=\fontfamily{phv}\selectfont\small\bfseries,
            anchor=north west,
            xshift=0.05cm,
            yshift=-0.05cm,
            at={(0,1)}
        }
    }
}

\newcommand{\rebuttal}[1]{{#1}}

\newcommand{\ie}{\textit{i}.\textit{e}., }

\newcommand{\secref}[1]{Sec.~\ref{#1}}

\newcommand{\figref}[1]{Fig.~\ref{#1}}
\newcommand{\tabref}[1]{Table~\ref{#1}}

\newcommand{\nvidia}{\textsc{Nvidia}\xspace}
\newcommand{\simName}{\textsc{Orbit}\xspace}

\title{\LARGE \bf
\simName: A Unified Simulation Framework for \\ Interactive Robot Learning Environments
}

\author{%
  Mayank Mittal$^{1, 2}$,
  Calvin Yu$^{3}$,
  Qinxi Yu$^{3}$,
  Jingzhou Liu$^{3}$,
  Nikita Rudin$^{1, 2}$,
  David Hoeller$^{1, 2}$,\\
  Jia Lin Yuan$^{3}$,
  Ritvik Singh$^{3}$,
  Yunrong Guo$^{2}$,
  Hammad Mazhar$^{2}$,
  Ajay Mandlekar$^{2}$,
  Buck Babich$^{2}$,\\
  Gavriel State$^{2}$,
  Marco Hutter$^{1}$,
  Animesh Garg$^{2, 3}$
\thanks{$^1$ ETH Zurich, Switzerland, $^2$ NVIDIA, $^3$ University of Toronto, Canada} %
\thanks{Corresponding author: Mayank Mittal (email: {\footnotesize \href{mailto:mittalma@ethz.ch}{mittalma@ethz.ch})}} %
}

\renewcommand{\baselinestretch}{0.965}

\usepackage{etoolbox}

\makeatletter
    \apptocmd{\@maketitle}{
        \centering
        \captionsetup{type=figure}
        \includegraphics[width=0.99\textwidth]{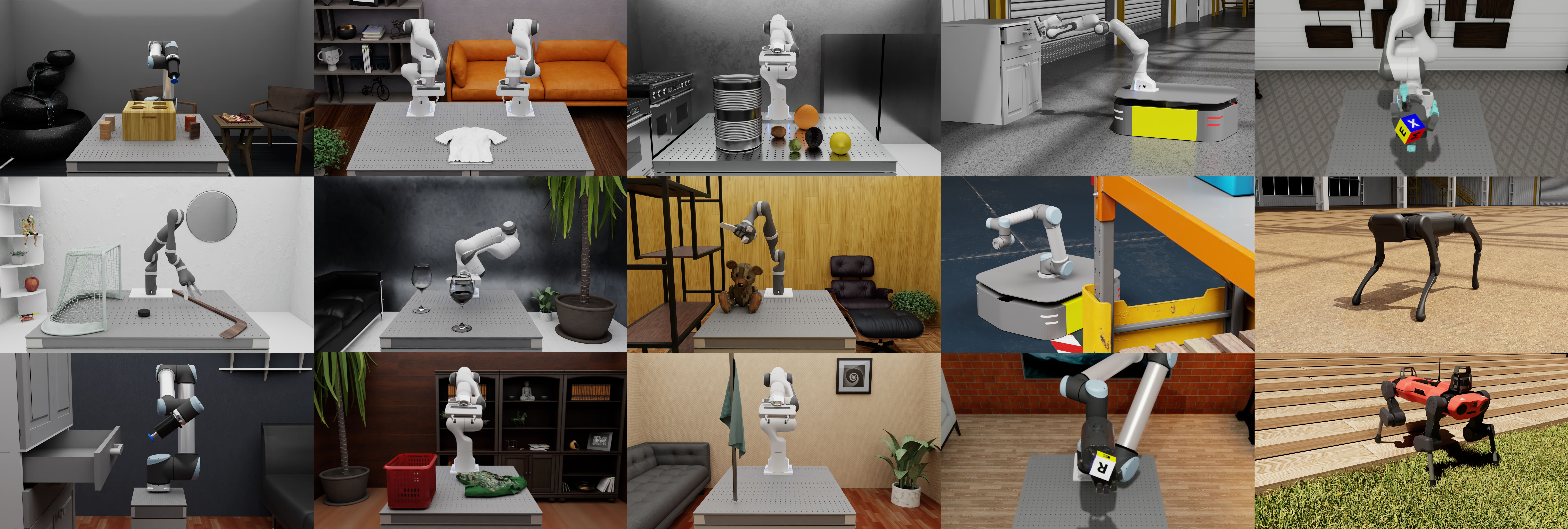}
        \captionof{figure}{\simName framework provides a large set of robots, sensors, rigid and deformable objects, motion generators, and teleoperation interfaces. Through these, we aim to simplify the process of defining new and complex environments, thereby providing a common platform for algorithmic research in robotics and robot learning.}
        \label{fig:teaser}
        \vspace{-20.5pt}
    }{}{}% insert the figure after authors
\makeatother
\begin{document}

\bstctlcite{IEEEexample:BSTcontrol}

\maketitle

\input{sections/0-abstract}

\input{sections/1-intro}

\input{sections/2-related-work}

\input{sections/3-interface}
\input{sections/4-features}

\input{sections/5-evaluation}
\input{sections/6-conclusion}

\section*{Acknowledgment}
We thank the \nvidia PhysX and Omniverse teams for their assistance with the simulator, as well as Farbod Farshidian and Mike Y. Michelis for their contributions to OCS2 and data collection of the deformable beam, respectively.

\bibliographystyle{bibtex/myIEEEtran} 
\bibliography{bibtex/IEEEabrv,bibtex/root}

\input{sections/7-appendix}

\end{document}

%% file: sections/0-abstract.tex
\begin{abstract}
  We present \simName, a unified and modular framework for robot learning powered by~\nvidia Isaac Sim.
  It offers a modular design to easily and efficiently create robotic environments with photo-realistic scenes and high-fidelity rigid and deformable body simulation. 
  With \simName, we provide a suite of benchmark tasks of varying difficulty-- from single-stage cabinet opening and cloth folding to multi-stage tasks such as room reorganization.
  To support working with diverse observations and action spaces, we include fixed-arm and mobile manipulators with different physically-based sensors and motion generators.
 \simName allows training reinforcement learning policies and collecting large demonstration datasets from hand-crafted or expert solutions in a matter of minutes by leveraging GPU-based parallelization.
  In summary, we offer an open-sourced framework that readily comes with 16 robotic platforms, 4 sensor modalities, 10 motion generators, more than 20 benchmark tasks, and wrappers to 4 learning libraries.
  With this framework, we aim to support various research areas, including representation learning, reinforcement learning, imitation learning, and task and motion planning.
  We hope it helps establish interdisciplinary collaborations in these communities, and its modularity makes it easily extensible for more tasks and applications in the future.
  For videos, documentation, and code: \url{https://isaac-orbit.github.io/}.
\end{abstract}

% \begin{IEEEkeywords}
% IEEE, IEEEtran, journal, \LaTeX, paper, template.
% \end{IEEEkeywords}
% \begin{IEEEkeywords}
% \lowercase{Software Tools for Benchmarking and Reproducibility, Machine Learning for Robot Control, Deep Learning for Visual Perception, Simulation and Animation}
% \end{IEEEkeywords}
\vspace{-4.5pt}

% \fi
\IEEEpeerreviewmaketitle

%% file: sections/1-intro.tex
\section{Introduction}

\input{sections/1-simulators-small}

An ideal robot simulator needs to provide fast and accurate physics, high-fidelity sensor simulation, diverse asset handling, and easy-to-use interfaces for integrating new tasks and environments. However, existing platforms often need to make a trade-off between these aspects.
For instance, simulators designed mainly for vision, such as Habitat~\cite{habitat19iccv} or ManipulaTHOR~\cite{Ehsani2021ManipulaTHORA}, offer decent rendering throughput but simplify low-level interaction intricacies such as grasping. On the other hand, physics simulators for robotics, such as Isaac Gym~\cite{makoviychuk2021isaac} or SAPIEN~\cite{xiang2020sapien}, provide fast and reasonably accurate rigid-body contact dynamics but do not include physically-based rendering (PBR), deformable objects simulation or ROS support~\cite{quigley2009ros} out-of-the-box. 
\rebuttal{Recently, \nvidia released a new simulator Omniverse Isaac Sim~\cite{nvidia2022isaacsim} that aims to fulfill these gaps through GPU-accelerated real-time PBR and state-of-the-art physics engine.}

This work presents \simName, an open-source framework for robotics research that exploits the latest simulation capabilities through Isaac Sim to allow intuitive designing of tasks with photo-realistic scenes and state-of-the-art rigid and deformable body simulation.
To prevent a scattering of efforts for building the necessary tooling to use the simulator for robot learning, we design a unified and modular framework that supports a wide variety of robotic platforms, sensors, and objects and allows designing tasks not only programmatically but also interactively through the GUI.
We design the system bottom-up -- from incorporating user-defined models for the actuator dynamics to modularizing task specifications for learning with different levels of observations and action spaces.
In \simName, we also include a collection of features, such as different robot platforms, motion generators, and a suite of tasks that help serve not only as a benchmark but also as examples for designing a new task.
Through this framework, we support various robotic applications, such as reinforcement learning (RL), learning from demonstrations (LfD), and motion planning, thereby providing a common platform for researchers in these communities to benefit from this consolidated effort.

Our main contributions are as follows:
\begin{enumerate}[noitemsep,]
    \item We design a unified and modular open-source framework for fast and flexible development, that leverages the latest advances in simulators for photo-realistic scenes and high-fidelity physics.
    \item It provides a batteries-included experience for roboticists with models readily available for different robots and sensors. This helps reduce the entry barrier to using the framework and its features.
    \item We include a suite of standardized tasks for benchmark purposes. These include eleven rigid object manipulation, thirteen deformable object manipulation, and two locomotion environments. Within each task, we allow switching robots, objects, and sensors easily.
    \item Through experiments, we demonstrate the accuracy of the simulator for rigid and soft body simulation. Compared to existing frameworks, we show that \simName is able to obtain up to ${\sim}10$x and ${\sim}3$x the throughput for rigid and deformable body manipulation tasks respectively. Additionally, we demonstrate the sim-to-real transfer of a locomotion policy for the quadruped robot, ANYmal.
\end{enumerate}

In the remainder of the paper, we describe the available simulator choices (\secref{sec:isaacsim}), the framework's design decisions and abstractions (\secref{sec:orbit-interfaces}), and its highlighted features (\secref{sec:features}). We demonstrate the framework's  applicability on different robotics workflows (\secref{sec:workflows}), show sim-to-real experiments for locomotion and manipulation, and evaluate the obtained accuracy and simulation throughput in \secref{sec:exp-sim-comp}.

%% file: sections/1-simulators-small.tex
\begin{table*}[!t]
\caption{Comparison between different simulation frameworks and \simName. The check ({\color[HTML]{00B323} \checkmark}) and cross ({\color[HTML]{FF0000} X}) denote presence or absence of the feature. In \textbf{Robotic Platforms} column, $M$ stands for manipulator. In \textbf{Scene Authoring} column, $G$ stands for game-based designing, $M$ for mesh-scan scenes, and $P$ for procedural-generation.}
\label{tab:benchmarks}
\begin{minipage}{1.0\textwidth}
\resizebox{\linewidth}{!}{%
\begin{threeparttable}
\begin{tabular}{l|ll|cc|cccc|c|ccccc|ccc|l}
\toprule
\multicolumn{1}{c}{{\color[HTML]{000000} }}                                & \multicolumn{1}{c}{{\color[HTML]{000000} }}                                                                                    & \multicolumn{1}{c}{{\color[HTML]{000000} }}                                                                                      & \multicolumn{2}{c}{{\color[HTML]{000000} \textbf{Vectorization}}}                                                 & \multicolumn{5}{c}{{\color[HTML]{000000} \textbf{Supported Dynamics}}}                                                                                                                                                                        & \multicolumn{1}{c}{{\color[HTML]{000000} }}                                                                                  & \multicolumn{4}{c}{{\color[HTML]{000000} \textbf{Sensors}}}                                                                                                                                                                                       & \multicolumn{3}{c}{{\color[HTML]{000000} \textbf{Robotic Platforms}}}                                                                                       
& \multicolumn{1}{c}{{\color[HTML]{000000} }}                                                                                     \\
\multicolumn{1}{c}{\multirow{-2}{*}{{\color[HTML]{000000} \textbf{Name}}}} & \multicolumn{1}{c}{\multirow{-2}{*}{{\color[HTML]{000000} \textbf{\begin{tabular}[c]{@{}c@{}}Physics Engine\end{tabular}}}}} & \multicolumn{1}{c}{\multirow{-2}{*}{{\color[HTML]{000000} \textbf{\begin{tabular}[c]{@{}c@{}}Renderer \end{tabular}}}}} & \multicolumn{1}{c}{{\color[HTML]{000000} \textbf{CPU}}} & \multicolumn{1}{c}{{\color[HTML]{000000} \textbf{GPU}}} & \multicolumn{1}{c}{{\color[HTML]{000000} \textbf{Rigid}}} & \multicolumn{1}{c}{{\color[HTML]{000000} \textbf{Cloth}}} & \multicolumn{1}{c}{{\color[HTML]{000000} \textbf{Soft}}} & \multicolumn{1}{c}{{\color[HTML]{000000} \textbf{Fluid}}} & \multicolumn{1}{c}{\multirow{-2}{*}{{\color[HTML]{000000} \textbf{\begin{tabular}[c]{@{}c@{}}PBR\\ Tracing\end{tabular}}}}} & \multicolumn{1}{c}{{\color[HTML]{000000} \textbf{RGBD}}} & \multicolumn{1}{c}{{\color[HTML]{000000} \textbf{Semantic}}} & \multicolumn{1}{c}{{\color[HTML]{000000} \textbf{LiDAR}}} & \multicolumn{1}{c}{{\color[HTML]{000000} \textbf{Contact}}} & \multicolumn{1}{c}{\rebuttal{\textbf{Acoustic}}} & \multicolumn{1}{c}{{\color[HTML]{000000} \textbf{Fixed-M}}} & \multicolumn{1}{c}{{\color[HTML]{000000} \textbf{Mobile-M}}} & \multicolumn{1}{c}{{\color[HTML]{000000} \textbf{Legged}}} & \multicolumn{1}{c}{\multirow{-2}{*}{{\color[HTML]{000000} \textbf{\begin{tabular}[c]{@{}c@{}}Scene\\ Authoring\end{tabular}}}}} 
\\ \midrule\midrule
{\color[HTML]{000000} MetaWorld~\cite {yu2020meta}}                   & {\color[HTML]{000000} MuJoCo}                                                                                                  & {\color[HTML]{000000} OpenGL}                                                                                                    & {\color[HTML]{00B323} \checkmark}                                & {\color[HTML]{FF0000} X}                                & {\color[HTML]{00B323} \checkmark}                                  & {\color[HTML]{FF0000} X}                                  & {\color[HTML]{FF0000} X}                                 & {\color[HTML]{FF0000} X}                                   & {\color[HTML]{FF0000} X}                                                                                                     & {\color[HTML]{FF0000} X}                                 & {\color[HTML]{FF0000} X}                                     & {\color[HTML]{FF0000} X}                                  & {\color[HTML]{FF0000} X}                                    & {\color[HTML]{FF0000} X} & {\color[HTML]{00B323} \checkmark}                                   & {\color[HTML]{FF0000} X}                                   & {\color[HTML]{FF0000} X}                                                   & {\color[HTML]{000000} $P$}                                                                                                        \\
{\color[HTML]{000000} RoboSuite~\cite {zhu2020robosuite}}                                           & {\color[HTML]{000000} MuJoCo}                                                                                                                          & {\color[HTML]{000000} OpenGL, OptiX}                                                                                                                     & {\color[HTML]{00B323} \checkmark}                                & {\color[HTML]{FF0000} X}                                & {\color[HTML]{00B323} \checkmark}                                  & {\color[HTML]{FF0000} X}                                  & {\color[HTML]{FF0000} X}                                 & {\color[HTML]{FF0000} X}                                   & {\color[HTML]{FF0000} X}                                                                                                     & {\color[HTML]{00B323} \checkmark}                                 & {\color[HTML]{00B323} \checkmark}                                     & {\color[HTML]{FF0000} X}                                  & {\color[HTML]{00B323} \checkmark}                                    & {\color[HTML]{FF0000} X} & {\color[HTML]{00B323} \checkmark}                                  & {\color[HTML]{FF0000} X}                                   & {\color[HTML]{FF0000} X}                                                                                 & {\color[HTML]{000000} $P$}                                                                                                        \\
{\color[HTML]{000000} DoorGym~\cite {DBLP:journals/corr/abs-1908-01887}}                                             & {\color[HTML]{000000} MuJoCo}                                                                                                                          & {\color[HTML]{000000} Unity}                                                                                                                             & {\color[HTML]{FF0000} X}                                & {\color[HTML]{FF0000} X}                                & {\color[HTML]{00B323} \checkmark}                                  & {\color[HTML]{FF0000} X}                                  & {\color[HTML]{FF0000} X}                                 & {\color[HTML]{FF0000} X}                                   & {\color[HTML]{00B323} \checkmark}                                                                                                     & {\color[HTML]{00B323} \checkmark}                                 & {\color[HTML]{00B323} \checkmark}                                     & {\color[HTML]{FF0000} X}                                  & {\color[HTML]{FF0000} X}                                    & {\color[HTML]{FF0000} X} & {\color[HTML]{00B323} \checkmark}                                  & {\color[HTML]{FF0000} X}                                   & {\color[HTML]{FF0000} X}                                                                                  & {\color[HTML]{000000} $P,G$}                                                                                                      \\
{\color[HTML]{000000} DEDO~\cite {antonova2021dedo}}                                                & {\color[HTML]{000000} Bullet}                                                                                                                          & {\color[HTML]{000000} OpenGL}                                                                                                                            & {\color[HTML]{00B323} \checkmark}                                & {\color[HTML]{FF0000} X}                                & {\color[HTML]{00B323} \checkmark}                                  & {\color[HTML]{00B323} \checkmark}                                  & {\color[HTML]{00B323} \checkmark}                                 & {\color[HTML]{FF0000} X}                                   & {\color[HTML]{FF0000} X}                                                                                                     & {\color[HTML]{00B323} \checkmark}                                 & {\color[HTML]{FF0000} X}                                     & {\color[HTML]{FF0000} X}                                  & {\color[HTML]{FF0000} X}                                    & {\color[HTML]{FF0000} X} & {\color[HTML]{00B323} \checkmark}                                  & {\color[HTML]{FF0000} X}                                   & {\color[HTML]{FF0000} X}                                                                              & {\color[HTML]{000000} $P,G$}                                                                                                      \\
{\color[HTML]{000000} RLBench~\cite {james2019rlbench}}                                             & {\color[HTML]{000000} Bullet/ODE}                                                                                                                      & {\color[HTML]{000000} OpenGL}                                                                                                                            & {\color[HTML]{FF0000} X}                                & {\color[HTML]{FF0000} X}                                & {\color[HTML]{00B323} \checkmark}                                  & {\color[HTML]{FF0000} X}                                  & {\color[HTML]{FF0000} X}                                 & {\color[HTML]{FF0000} X}                                   & {\color[HTML]{FF0000} X}                                                                                                     & {\color[HTML]{00B323} \checkmark}                                 & {\color[HTML]{00B323} \checkmark}                                     & {\color[HTML]{FF0000} X}                                  & {\color[HTML]{00B323} \checkmark}                                    & {\color[HTML]{FF0000} X} & {\color[HTML]{00B323} \checkmark}                                  & {\color[HTML]{FF0000} X}                                   & {\color[HTML]{FF0000} X}                                                  & {\color[HTML]{000000} $P$}                                                                                                        \\
{\color[HTML]{000000} iGibson~\cite {li2021igibson}}                                             & {\color[HTML]{000000} Bullet}                                                                                                                          & {\color[HTML]{000000} MeshRenderer}                                                                                                                      & {\color[HTML]{00B323} \checkmark}                                & {\color[HTML]{FF0000} X}                                & {\color[HTML]{00B323} \checkmark}                                  & {\color[HTML]{FF0000} X}                                  & {\color[HTML]{FF0000} X}                                 & {\color[HTML]{FF0000} X}                                   & {\color[HTML]{00B323} \checkmark}                                                                                                     & {\color[HTML]{00B323} \checkmark}                                 & {\color[HTML]{00B323} \checkmark}                                     & {\color[HTML]{00B323} \checkmark}                                  & {\color[HTML]{FF0000} X}                                    & {\color[HTML]{FF0000} X} & {\color[HTML]{FF0000} X}                                  & {\color[HTML]{00B323} \checkmark}                                   & {\color[HTML]{FF0000} X}                                                                            & {\color[HTML]{000000} $M$}                                                                                                        \\
{\color[HTML]{000000} Habitat 2.0~\cite {szot2021habitat2}}                                         & {\color[HTML]{000000} Bullet}                                                                                                                          & {\color[HTML]{000000} Magnum}                                                                                                                            & {\color[HTML]{FF0000} X}                                & {\color[HTML]{FF0000} X}                                & {\color[HTML]{00B323} \checkmark}                                  & {\color[HTML]{FF0000} X}                                  & {\color[HTML]{FF0000} X}                                 & {\color[HTML]{FF0000} X}                                   & {\color[HTML]{00B323} \checkmark}                                                                                                     & {\color[HTML]{00B323} \checkmark}                                 & {\color[HTML]{00B323} \checkmark}                                     & {\color[HTML]{FF0000} X}                                  & {\color[HTML]{FF0000} X}                                    & {\color[HTML]{FF0000} X} & {\color[HTML]{00B323} \checkmark}                                  & {\color[HTML]{00B323} \checkmark}                                   & {\color[HTML]{00B323} \checkmark}                                                                      & {\color[HTML]{000000} $P,M$}                                                                                                      \\
{\color[HTML]{000000} SoftGym~\cite {Lin2020softgym}}                                             & {\color[HTML]{000000} FleX}                                                                                                                            & {\color[HTML]{000000} OpenGL}                                                                                                                            & {\color[HTML]{00B323} \checkmark}                                & {\color[HTML]{FF0000} X}                                & {\color[HTML]{00B323} \checkmark}                                  & {\color[HTML]{00B323} \checkmark}                                  & {\color[HTML]{00B323} \checkmark}                                 & {\color[HTML]{00B323} \checkmark}                                   & {\color[HTML]{FF0000} X}                                                                                                     & {\color[HTML]{FF0000} X}                                 & {\color[HTML]{FF0000} X}                                     & {\color[HTML]{FF0000} X}                                  & {\color[HTML]{FF0000} X}                                    & {\color[HTML]{FF0000} X} & {\color[HTML]{00B323} \checkmark}                                  & {\color[HTML]{FF0000} X}                                   & {\color[HTML]{FF0000} X}                                                                       & {\color[HTML]{000000} $P$}                                                                                                        \\
{\color[HTML]{000000} ThreeDWorld~\cite {gan2021threedworld}}                                         & {\color[HTML]{000000} PhysX 4/FleX/Obi}                                                                                                                  & {\color[HTML]{000000} Unity3D}                                                                                                                           & {\color[HTML]{FF0000} X}                                & {\color[HTML]{FF0000} X}                                & {\color[HTML]{00B323} \checkmark}$^*$                                  & {\color[HTML]{00B323} \checkmark}$^*$                                 & {\color[HTML]{00B323} \checkmark}$^*$                                 & {\color[HTML]{00B323} \checkmark}$^*$                                   & {\color[HTML]{00B323} \checkmark}                                                                                                     & {\color[HTML]{00B323} \checkmark}                                 & {\color[HTML]{00B323} \checkmark}                                     & {\color[HTML]{FF0000} X}                                  & {\color[HTML]{FF0000} X}                                    & {\color[HTML]{00B323} \checkmark} & {\color[HTML]{FF0000} X}                                  & {\color[HTML]{00B323} \checkmark}                                   & {\color[HTML]{FF0000} X}                                                                       & {\color[HTML]{000000} $P$}                                                                                                        \\
{\rebuttal{ManiSkill2~\cite {gu2023maniskill2}}}                                              & {\rebuttal{ PhysX 4/Warp}}                                                                                                                         & {\rebuttal{SAPIEN~\cite{xiang2020sapien}}}                                                                                                                      & {\color[HTML]{00B323} \checkmark}                                & {\color[HTML]{FF0000} X}                                & {\color[HTML]{00B323} \checkmark}                                  & {\color[HTML]{FF0000} X}                                  & {\color[HTML]{00B323} \checkmark}$^\dagger$                                 & {\color[HTML]{00B323} \checkmark}                                   & {\color[HTML]{00B323} \checkmark}                                                                                                     & {\color[HTML]{00B323} \checkmark}                                 & {\color[HTML]{00B323} \checkmark}                                     & {\color[HTML]{FF0000} X}                                  & {\color[HTML]{00B323} \checkmark}                                    & {\color[HTML]{FF0000} X} & {\color[HTML]{00B323} \checkmark}                                  & {\color[HTML]{00B323} \checkmark}                                   & {\color[HTML]{FF0000} X}                                                                                 & {\color[HTML]{000000} $P$}                                                                                                        \\
{\color[HTML]{000000} ManipulatorThor~\cite {Ehsani2021ManipulaTHORA}}                                     & {\color[HTML]{000000} PhysX 4}                                                                                                                         & {\color[HTML]{000000} Unity}                                                                                                                             & {\color[HTML]{FF0000} X}                                & {\color[HTML]{FF0000} X}                                & {\color[HTML]{00B323} \checkmark}                                  & {\color[HTML]{FF0000} X}                                  & {\color[HTML]{FF0000} X}                                 & {\color[HTML]{FF0000} X}                                   & {\color[HTML]{00B323} \checkmark}                                                                                                     & {\color[HTML]{00B323} \checkmark}                                 & {\color[HTML]{00B323} \checkmark}                                     & {\color[HTML]{FF0000} X}                                  & {\color[HTML]{FF0000} X}                                    & {\color[HTML]{FF0000} X} & {\color[HTML]{FF0000} X}                                  & {\color[HTML]{00B323} \checkmark}                                   & {\color[HTML]{FF0000} X}                                                                                & {\color[HTML]{000000} $P,G$}                                                                                                      \\
{\color[HTML]{000000} IsaacGymEnvs~\cite {makoviychuk2021isaac}}                                        & {\color[HTML]{000000} PhysX 5}                                                                                                                         & {\color[HTML]{000000} Vulkan}                                                                                                                            & {\color[HTML]{00B323} \checkmark}                                & {\color[HTML]{00B323} \checkmark}                                & {\color[HTML]{00B323} \checkmark}                                  & {\color[HTML]{FF0000} X}                                  & {\color[HTML]{FF0000} X}                                 & {\color[HTML]{FF0000} X}                                   & {\color[HTML]{FF0000} X}                                                                                                     & {\color[HTML]{00B323} \checkmark}                                 & {\color[HTML]{00B323} \checkmark}                                     & {\color[HTML]{FF0000} X}                                  & {\color[HTML]{00B323} \checkmark}                                    & {\color[HTML]{FF0000} X} & {\color[HTML]{00B323} \checkmark}                                  & {\color[HTML]{FF0000} X}                                   & {\color[HTML]{00B323} \checkmark}                                                                          & {\color[HTML]{000000} $P$}                                                                                                        \\ \midrule
{\color[HTML]{000000} \textbf{\simName (ours)}}                                               & {\color[HTML]{000000} PhysX 5.1}                                                                                                                         & {\color[HTML]{000000} Omniverse RTX}                                                                                                                      & {\color[HTML]{00B323} \checkmark}                                & {\color[HTML]{00B323} \checkmark}                                & {\color[HTML]{00B323} \checkmark}                                  & {\color[HTML]{00B323} \checkmark}                                  & {\color[HTML]{00B323} \checkmark}$^\dagger$                                 & {\color[HTML]{00B323} \checkmark}                                   & {\color[HTML]{00B323} \checkmark}                                                                                                     & {\color[HTML]{00B323} \checkmark}                                 & {\color[HTML]{00B323} \checkmark}                                     & {\color[HTML]{00B323} \checkmark}                                  & {\color[HTML]{00B323} \checkmark}                                    & {\color[HTML]{FF0000} X} & {\color[HTML]{00B323} \checkmark}                                  & {\color[HTML]{00B323} \checkmark}                                   & {\color[HTML]{00B323} \checkmark}                                                                       & {\color[HTML]{000000} $P,M,G$}\\\bottomrule                                                                                                   
\end{tabular}
\vspace{2pt}
\begin{tablenotes}\footnotesize
\item[*] ThreeDWorld supports simulation of rigid bodies and deformable bodies based on whether PhysX 4 or FleX/Obi is enabled respectively. Thus, it is limited in simulating interactions between rigid and deformable bodies.
\item[$\dagger$] ManiSkill2 supports a Warp-based Material Point Method (MPM) solver that helps simulate cutting and plastic deformations of soft objects. Currently, this feature is under development for \simName.
\end{tablenotes}
\end{threeparttable}
}
\end{minipage}
\vspace{-10pt}
\end{table*}

%% file: sections/2-related-work.tex
\section{Related Work}
\label{sec:isaacsim}

Recent years have seen several simulation frameworks, each specializing in particular robotic applications. In this section, we highlight the design choices crucial for building a unified simulation platform and how \simName compares to other frameworks~(also summarized in~\tabref{tab:benchmarks}).

\paragraph{Physics Engine} 
Increasing the complexity and realism of physically simulated environments is essential for advancing robotics research. This includes improving the contact dynamics, having better collision handling for non-convex geometries (such as threads), stable solvers for deformable bodies, and high simulation throughput.

Prior frameworks~\cite{zhu2020robosuite,james2019rlbench} using MuJoCo~\cite{todorov2012mujoco} or Bullet~\cite{coumans2021bullet} focus mainly on rigid object manipulation tasks. Since their underlying physics engines are CPU-based, they need CPU clusters to achieve massive parallelization~\cite{makoviychuk2021isaac}. On the other hand, frameworks for deformable bodies~\cite{antonova2021dedo, Lin2020softgym} mainly employ Bullet~\cite{coumans2021bullet} or FleX~\cite{flex}, which use particle-based dynamics for soft bodies and cloth simulation.
\rebuttal{Recently, SofaGym~\cite{menager2022sofagym}, an RL framework specifically for soft robotics, shows the benefits of using finite-element-methods (FEM) for deformation models.
}.
However, limited tooling exists in these frameworks compared to those for rigid object tasks. \simName aims to bridge this gap by providing an integrated robotics framework that supports rigid and deformable body simulation via PhysX SDK 5~\cite{nvidia2022physx}. In contrast to other engines, PhysX SDK 5 features GPU-based hardware acceleration for high throughput, signed-distance field (SDF) collision checking~\cite{narang2022factory}, and more stable solvers based on FEM for deformable body simulation~\cite{neohooken2021miles, xpbd2019miles}.

\paragraph{Sensor simulation} 
Various existing frameworks~\cite{zhu2020robosuite,james2019rlbench,makoviychuk2021isaac} use classic rasterization that limits the photo-realism in the generated images. Recent techniques~\cite{parker2010optix} simulate the interaction of rays with object's textures in a physically correct manner. These methods help capture fine visual properties such as transparency and reflection, thereby promising for bridging the sim-to-real visual domain gap.
While recent frameworks~\cite{xiang2020sapien,szot2021habitat2,Ehsani2021ManipulaTHORA} include physically-based renderers, they mainly support camera-based sensors (RGB, depth). This is insufficient for certain mobile robot applications that need range sensors, such as \href{https://developer.nvidia.com/blog/validating-active-sensors-in-nvidia-drive-sim/}{LiDARs}. Leveraging the ray-tracing technology in \nvidia Isaac Sim, \simName supports all these modalities and includes APIs to obtain additional information such as semantic annotations. %

\begin{figure*}
    \centering
    \includegraphics[width=0.8\textwidth]{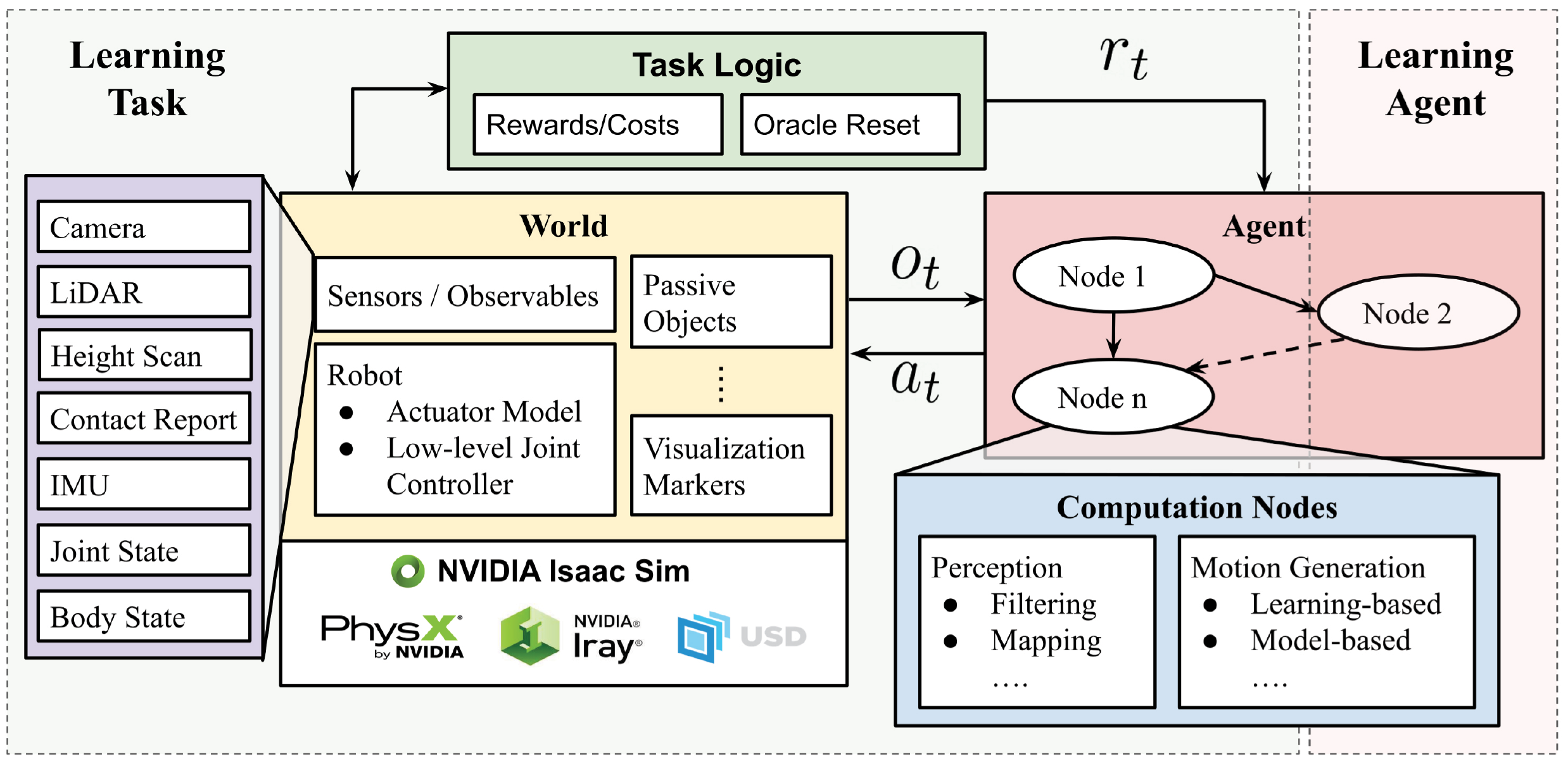}
    \caption{\simName's abstractions comprise \texttt{World}, analogous to the real world, and \texttt{Agent}, the computation graph behind the embodied system. The nodes in the agent's graph can perform observation-based or action-based processing. Through a graph-cut over this computation graph and specifying an extrinsic goal, it is feasible to design different tasks within the same \texttt{World} definition. \rebuttal{For instance, to perform RL, this would define the task with $o_t$, $a_t$, and $r_t$ corresponding to the observation, action, and reward signals respectively.}}
    \label{fig:design}
    \vspace{-12pt}
\end{figure*}

\paragraph{Scene designing and asset handling}
Frameworks support scene creation procedurally~\cite{yu2020meta,zhu2020robosuite,xiang2020sapien}, via mesh scans~\cite{li2021igibson, szot2021habitat2} or through game-engine style interfaces~\cite{airsim2017fsr,gan2021threedworld}. While mesh scans simplify generating large amounts of scenes, they often suffer from geometric artifacts and lighting problems. On the other hand, procedural generation allows leveraging object datasets for diverse scenes. \rebuttal{While Isaac Sim mainly focuses on GUI-based scene designing, we choose to not be restrictive and build upon its interfaces to support all three methods for scene design.}

%% file: sections/3-interface.tex
\section{\simName: Abstractions and Interfaces Design}
\label{sec:orbit-interfaces}

At a high level, the framework design comprises a \texttt{world} and an \texttt{agent}, similar to the real world and the software stack running on the robot. The agent receives raw observations from the world and computes the actions to apply to the embodiment (\texttt{robot}). Typically in learning, it is assumed that all the perception and motion generation occurs at the same frequency. However, in the real world, that is rarely the case: (1) sensors update at differing frequencies, (2) depending on the control architecture, actions are applied at different time-scales~\cite{martin2019vices}, and (3) unmodeled sources of delays and noises are present in the system. \rebuttal{In \simName, we thoughtfully design its interfaces and abstractions to support these functionalities and allow the inclusion of actuator and noise models to bridge the sim-to-real gap.}

\paragraph{\textbf{World}} Analogous to the real world, we define a \texttt{world} where \texttt{robots}, \texttt{sensors}, \texttt{objects} (static or dynamic), and \rebuttal{\texttt{visualization markers}} exist on the same stage. The \texttt{world} can be designed procedurally (script-based), via scanned meshes~\cite{straub2019replica}, or interactively through the game-based GUI of Isaac Sim, or a combination of them. This flexibility reaps the benefits of 3D reconstructed meshes, which capture various architectural layouts, with game-based designing, that simplifies the experience of creating and verifying the scene physics properties by playing the simulation.

\begin{figure}[!t]
    \centering
    \includegraphics[width=\linewidth]{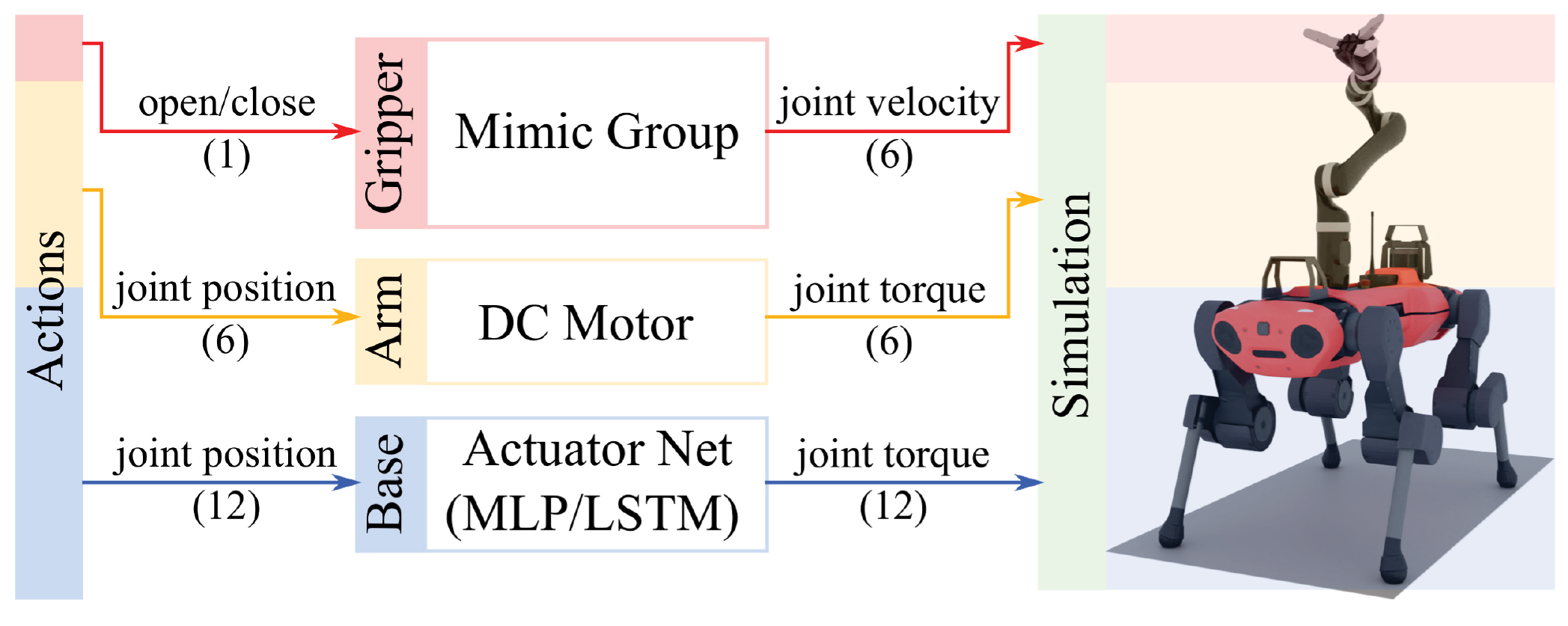}
    \caption{Illustration of actuator groups for a legged mobile manipulator. This allows decomposing a complex system into sub-groups and defining of specific transmission models for each of them flexibly. \rebuttal{The number inside $(\cdot)$ is the dimension of the command vector.}}
    \label{fig:actuator_group}
    \vspace{-15pt}
\end{figure}

\texttt{Robots} are a crucial component of the \texttt{world} since they serve as the embodiment for interaction. They consist of articulation, \rebuttal{actuator models}, and low-level \rebuttal{joint} controllers. \rebuttal{We design these interfaces such that a single interface (such as \texttt{LeggedRobot}) supports a variety of robots belonging to the same category (such as ANYmal C or Unitree A1). This simplifies setting up a new robot with the simulator and using it in an existing \texttt{World}.} The \texttt{robot} class loads the robot model from a USD file\footnote{\rebuttal{\href{https://www.nvidia.com/en-us/omniverse/usd/}{\underline{Universal Scene Description (USD)}} is a hierarchical file format, used in Omniverse, that allows storing assets, materials definitions, and various attributes (such as for physics, semantics, rendering) efficiently and flexibly. \simName includes scripts to convert assets of different formats (URDF, obj, stl) into USD from the command line and adds various physics properties to them automatically, such as colliders and friction materials. }}. \rebuttal{It creates and manages the necessary physics handles to set and read the simulation state.} \rebuttal{The actuator models play an important role in injecting real-world actuator characteristics into the simulation, such as delays or torque saturation. This can help facilitate the sim-to-real transfer of control policies on hardware (\secref{sec:real-robot-exp}). To apply actions on the robot, the joint-level controller processes input commands through actuator models before applying the desired joint position, velocity, or torque commands to the simulator (as shown in~\figref{fig:actuator_group}). Currently, we include actuator models for Direct Control (DC) motors and Series Elastic Actuators (SEA)~\cite{hwango2019actuator}. However, it is easy for users to integrate the actuator model for their robot into \simName after identifying its dynamics characteristics.}

\begin{figure*}[!t]
\centering
    \includegraphics[width=\textwidth]{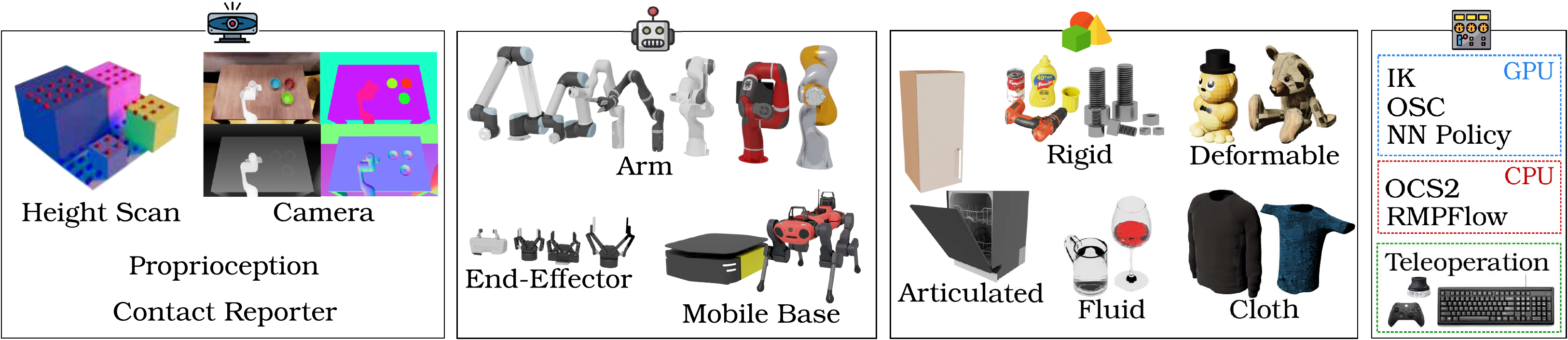}
    \caption{Overview of features included in \simName. We provide models of different sensors, robotic platforms, objects from different datasets, motion generators, and teleoperation devices. Using RTX-accelerated ray-tracing, we can obtain high-fidelity images in real-time for different modalities such as RGB, depth, surface normal, instance, and semantic segmentation (pixel-wise and bounding boxes).}
    \label{fig:features}
    \vspace{-12pt}
\end{figure*}

\texttt{Sensors} (proprioceptive or exteroceptive) may exist both for the \texttt{robot} or externally (such as third-person cameras). While \simName relies on Isaac Sim for different physics-based (range, force, and contact sensor) and rendering-based (RGB, depth, normals) sensors, it unifies them under a common interface \rebuttal{to simplify creating and configuring them into a scene at runtime. We diverge from the recommended USD practice in Isaac Sim where sensors are expected to be specified in a USD file and all of them are updated at every simulation step. While this practice doesn't cause a huge issue for the traditional robotics paradigm, it leads to significant overhead when having parallelized scenes as there would exist undesired sensors updating in the scene. Through a common sensor management system in the \texttt{world}, \simName configures the sensors required only for a given task.} To simulate sensing at different frequencies, each sensor instance has its own internal timer that governs the operating frequency for reading the simulator buffers. Between the timesteps, the sensor returns the previously obtained values.

\texttt{Objects} are passive entities in the \texttt{world}. While several objects may exist in the scene, the user can define objects of interest for a specified task and set and retrieve properties/states only for them.
For any given object, we support the randomization of its textures and physics properties, such as friction material and joint parameters.
\rebuttal{Similar to the robots class, we also include interfaces to augment the simulator with models for active motion components in an object. For instance, in a refrigerator, the hinge joint is stiff initially due to a magnetic seal but becomes free once the seal is broken. Accounting for these hybrid models in the simulation allows for developing and verifying methods to deal with the varying object dynamics in the real world.}

\rebuttal{Addition to above, the \texttt{world} also constitutes \texttt{visualization markers}. We allow programmatic additions of various primitive shapes to the simulator's GUI, such as axes, spheres, and meshes. These are often useful for development purposes, such as displaying the goal state or visualizing different coordinate frames. We integrate these markers into different robots and controller interfaces to readily allow visualization of useful frames such as the feet or end-effector poses.}

\paragraph{\textbf{Agent}} An \texttt{agent} refers to the decision-making process (``intelligence'') guiding the embodied system. While roboticists have embraced the modularity of ROS~\cite{quigley2009ros}, most robot learning frameworks often focus only on the environment (or MDP) definition. This practice leads to code replication and adds friction to switching between different implementations. 

Keeping modularity at its core, an agent in \simName comprises various nodes that formulate a computation graph exchanging information between them. Broadly, we consider nodes are of two types: 1) \textit{perception-based} \ie they process inputs into another representation (such as RGB-D image to point-cloud/TSDF), or 2) \textit{action-based} \ie they process inputs into action commands (such as task-level commands to joint commands). \rebuttal{Similar to sensors, nodes also contain their own internal timers that control their update frequency.} Currently, the flow of information between nodes happens synchronously via Python, which avoids the data exchange overhead of service-client protocols.

\paragraph{\textbf{Learning task and agent}} Paradigms such as RL require specifying a task, a \texttt{world} and may include some computation nodes of the \texttt{agent}.
The \texttt{task logic} helps specify the goal for the agent, compute metrics (rewards) to evaluate the agent's performance, and manage resets. With this component as a separate module, it becomes feasible to use the same \texttt{world} definition for different tasks, similar to learning in the real world, where tasks are specified through extrinsic reward signals. 
The \texttt{task} definition may also contain different nodes of the agent. An intuitive way to formalize this is by considering that learning for a particular node happens through a \textit{graph cut} on the agent's computation graph.

To further concretize the design motivation, consider the example of learning over task space instead of low-level joint actions for lifting a cube~\cite{martin2019vices}. In this case, the task-space controller, such as inverse kinematics (IK), would typically run at 50 Hz, while the joint controller requires commands at 1000 Hz. Although the task-space controller is a part of the agent's and not the world's computation, it is possible to encapsulate that into the \texttt{task} design. This functionality easily allows switching between motion generators, such as IK, operational space control (OSC), or reactive planners. %

\begin{figure*}[!t]
    \centering
    \includegraphics[width=\linewidth]{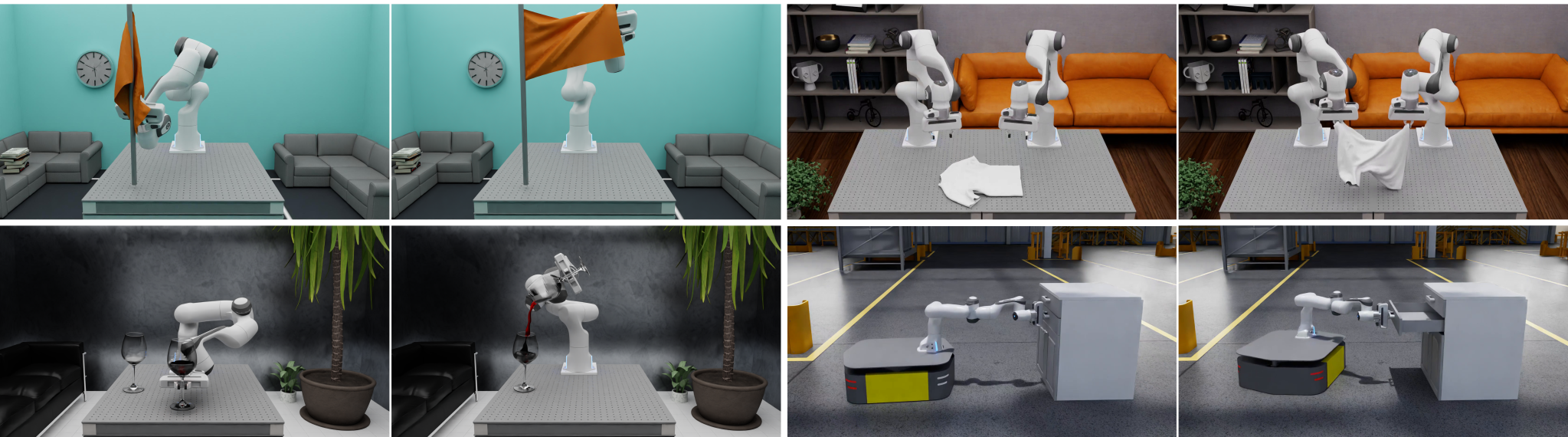}
    \caption{Demonstration of the designed tasks using hand-crafted state machines and task-space controllers. Leveraging recent advances in physics engines, we support high-fidelity simulation of rigid and deformable objects. We include environments that allow switching between robots, objects, observations, and action spaces through configuration files (\href{https://isaac-orbit.github.io/\#SampleTasks}{videos}). }
    \label{fig:cloth-manip}
    \vspace{-14pt}
\end{figure*}

%% file: sections/4-features.tex
\section{\simName: Features}
\label{sec:features}
While various robotic benchmarks have been proposed~\cite{antonova2021dedo,yu2020meta,james2019rlbench}, the right choice of necessary and sufficient tasks to demonstrate “intelligent” behaviors remains an open question. Instead of being prescriptive about tasks, we provide \simName as a platform to easily design new tasks. To facilitate the same, we include a diverse set of supported robots, peripheral devices, and motion generators and a large set of tasks for rigid and soft object manipulation for essential skills such as folding cloth, opening the dishwasher, and screwing a nut into a bolt. Each task showcases aspects of physics and renderer that we believe will facilitate answering crucial research questions, such as building representations for deformable object manipulation and learning skills that generalize to different objects and robots.

\paragraph{Robots}
We support 4 mobile platforms (one omnidirectional drive base and three quadrupeds), 7 robotic arms (two 6-DoF and five 7-DoF), and 6 end-effectors (four parallel-jaw grippers and two robotic hands). We provide tools to compose different combinations of these articulations into a complex robotic system such as a legged mobile manipulator. This provides a large set of robot platforms, each of which can be switched in the \texttt{World}.

\paragraph{I/O Devices}
Devices define the interface to peripheral controllers that teleoperate the robot in real-time. The interface reads the input commands from an I/O device and parses them into control commands for subsequent nodes. This helps not only in collecting demonstrations but also in debugging the task designs. Currently, we include support for Keyboard, Gamepad (Xbox controller), and Spacemouse.

\paragraph{Motion Generators}
Motion generators transform high-level actions into lower-level commands by treating input actions as reference tracking signals. For instance, inverse kinematics (IK)~\cite{ikJac} interprets commands as the desired end-effector poses and computes the desired joint positions. Employing these controllers, particularly in task space, has been shown to help sim-to-real transferability of robot manipulation policies~\cite{zhu2020robosuite,martin2019vices}.%

With \simName, we include GPU-based implementations for differential IK~\cite{ikJac}, operational-space control~\cite{khatib1995osc}, and joint-level control, \rebuttal{which compute commands for several robots efficiently}. Additionally, we provide CPU implementation of state-of-the-art model-based planners such as RMP-Flow~\cite{RMPFlow} for fixed-arm manipulators and OCS2~\cite{mittal2021articulated} for whole-body control of mobile manipulators. To facilitate research in legged robot navigation, \rebuttal{such as traversability estimation and path-planning}, we also include pre-trained policies for legged locomotion~\cite{rudin2022learning} that track base velocity commands. %

\paragraph{Rigid- and Deformable-body Tasks}
\rebuttal{\simName includes a suite of tasks for deformable and rigid object manipulation, as well as legged robot control and in-hand manipulation. These tasks serve not only as benchmarks for robotics research but also as examples for users to design new tasks easily. While some of these tasks have existed in prior works~\cite{yu2020meta,james2019rlbench,narang2022factory,rudin2022learning}, we enhance them using the framework's interfaces to simplify switching between different robots, objects, motion generators, observations, and domain randomization. We also extend manipulation tasks for fixed-arm robots to mobile manipulators. Currently, the tasks mainly focus on a diverse set of skills such as grasping, screwing, stacking, pushing/pulling, pouring, folding, and walking (shown in~\figref{fig:cloth-manip}). A complete and growing list of environments is available on our \href{https://isaac-orbit.github.io/orbit/source/features/environments.html}{\underline{website}}.}

%% file: sections/5-evaluation.tex
\section{Exemplar Workflows with \simName}
\label{sec:workflows}
\simName is a unified simulation infrastructure that provides both pre-built environments and easy-to-use interfaces that enables extendability and customization. Owing to high-quality physics, sensor simulation, and rendering, \simName is useful for multiple robotics challenges in both perception and decision-making. We outline a subset of such use cases through exemplar workflows. 

\subsection{Reinforcement Learning}

\rebuttal{Since the framework primarily stores data as tensors, the data needs to be formatted and converted to data types for different learning frameworks. We provide wrappers to rl-games~\cite{rl-games2022}, RSL-rl~\cite{rudin2022learning}, and stable-baselines-3~\cite{stable-baselines3}. These wrappers make the underlying environment RL framework agnostic and provide users access to a larger set of algorithms for research.}

\begin{figure}[!t]
    \centering
    \minipage{0.48\linewidth}
    \includegraphics[width=\linewidth]{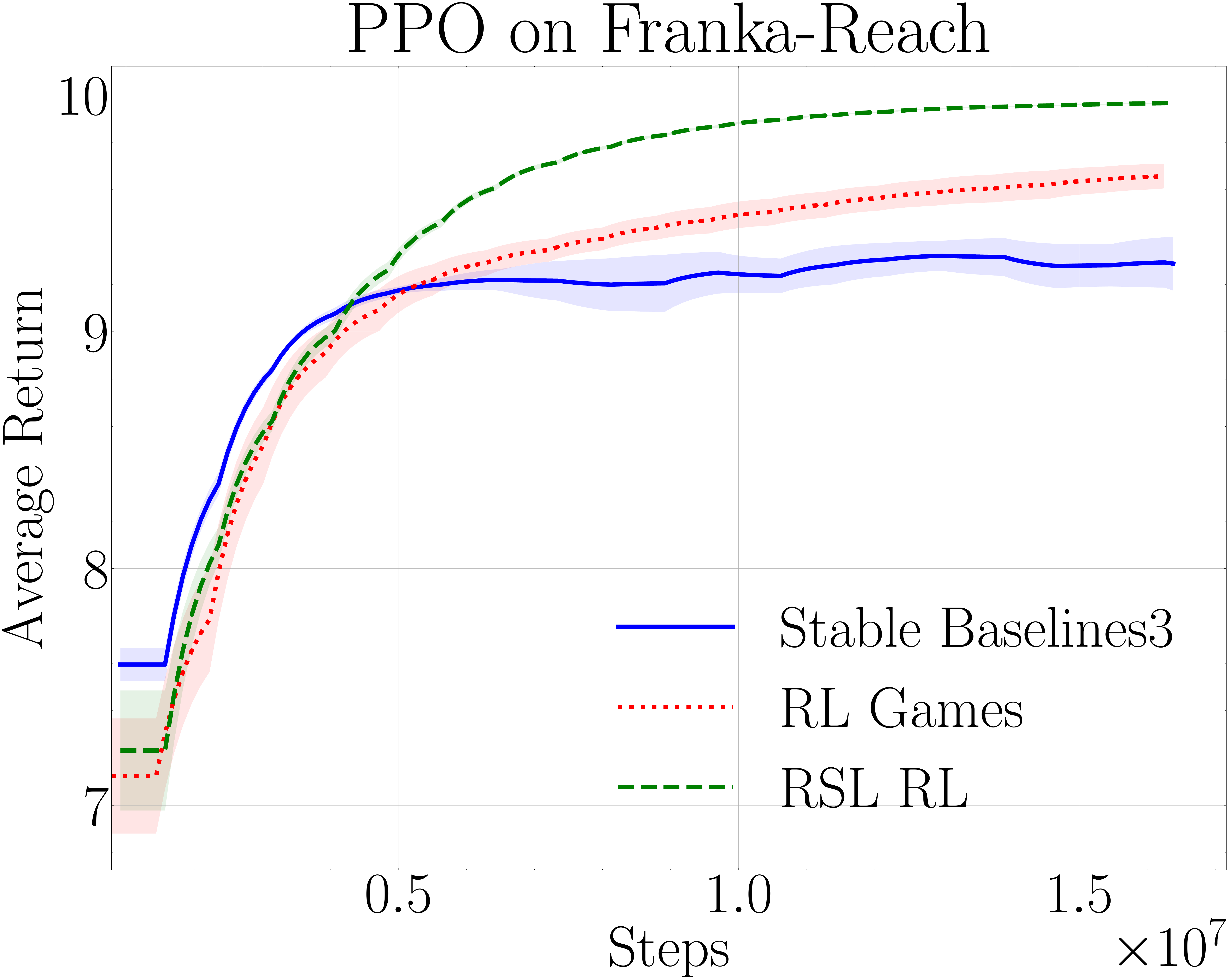}
    \endminipage
    \hfill     
    \minipage{0.48\linewidth}
    \includegraphics[width=\linewidth]{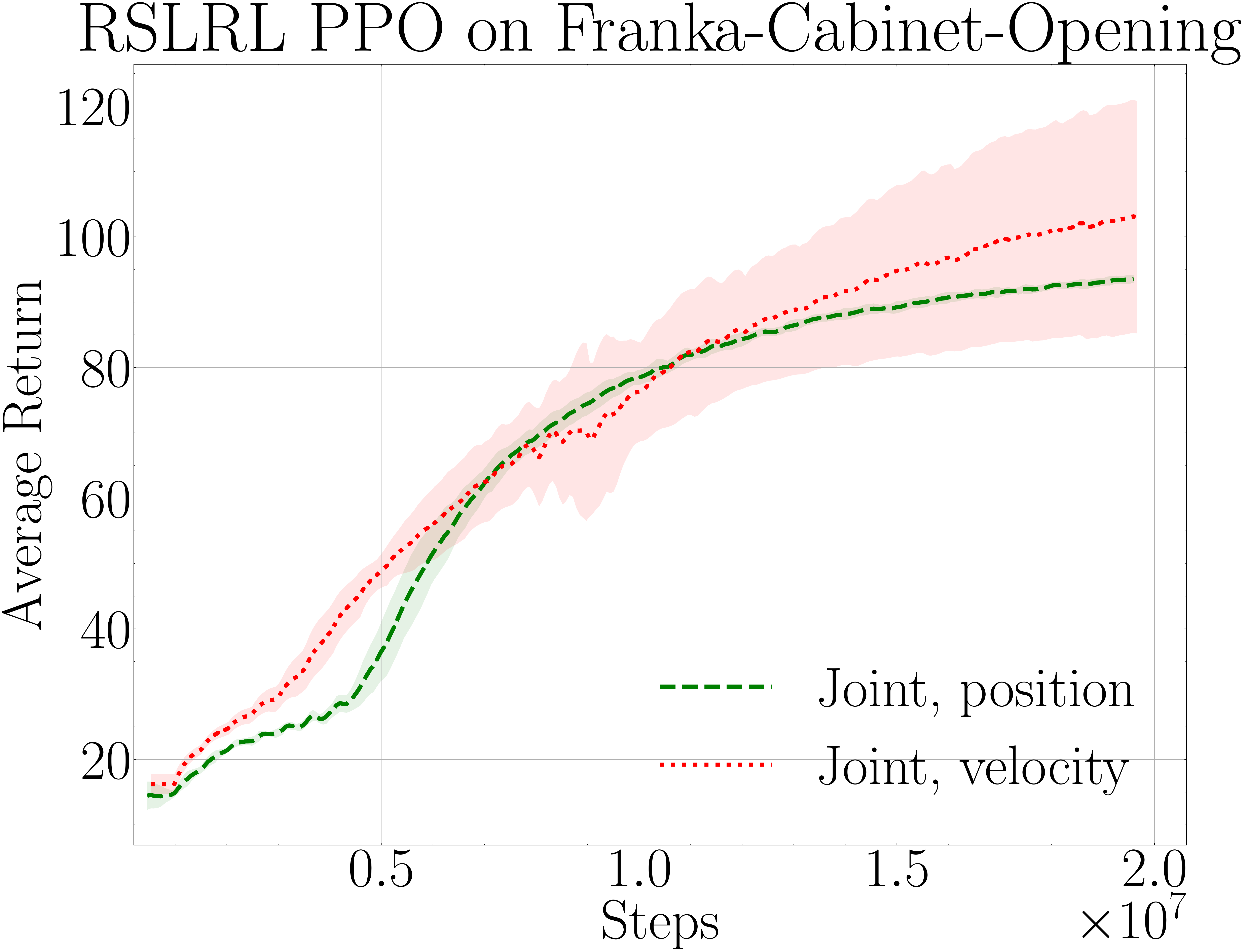}
    \endminipage
    \caption{\rebuttal{Example showing RL integration. We include wrappers to various RL frameworks. Additionally, it is possible to easily switch action spaces for training policies with different controllers.} 
    The plot shows the mean of the average return over five seeds.}
    \label{fig:rl-experiments}
    \vspace{-5pt}
\end{figure}

In \figref{fig:rl-experiments}, we show the training of \texttt{Franka-Reach} and \texttt{Franka-Cabinet-Opening} with PPO~\cite{schulman2017proximal} using different RL frameworks and action spaces. Although we ensure the same parameter settings for PPO in the frameworks, we notice a difference in their performance and training time due to implementation differences. Since RSL-rl and rl-games are optimized for GPU, we observe a training speed of 50,000-75,000 frames per second (FPS) with 2048 environments, while with stable-baselines3, we receive 6,000-18,000 FPS.

\subsection{Teleoperation and Imitation Learning}
Many manipulation tasks are computationally expensive or beyond the reach of current RL algorithms. 
In these scenarios, bootstrapping from user demonstrations provides a viable path to skill learning. 
\simName provides a data collection interface that is useful for interacting with environments using I/O devices and collecting data similar to roboturk~\cite{mandlekar2018roboturk}. We also support \rebuttal{storing the data in the format desired by robomimic~\cite{mandlekar2021robomimic}}, which provides access to training various imitation learning (IL) models through it.

\begin{table}[!t]
\centering
\begin{minipage}{0.47\textwidth}
\caption{\rebuttal{Showcase of collecting demonstrations using teleoperation and performing imitation learning with them.} We evaluate the trained policies for \texttt{Franka-LiftCube} in the same setting (\textbf{No Change}), changing initial states (\textbf{I)}, goal states (\textbf{G}), and changing both initial and goal states (\textbf{Both}). The success rate and trajectory lengths are reported over 100 trials.}
\label{tab:imitation}
\end{minipage}
\hfill
\begin{minipage}{0.5\textwidth}
\resizebox{\linewidth}{!}{%
\begin{tabular}{ c|c|c|c } 
\toprule
 \textbf{Algorithm} & \textbf{Avg. Traj. Len} &  \textbf{Succ. Rate} & \textbf{Eval. Setup} \\
 \midrule\midrule
 \multirow{4}{0.7in}{\textbf{BC}\ $\vert$\ \textbf{BC-RNN} } &  234 $\vert$ ~249 & 1.00 $\vert$ ~1.00 & \textbf{No Change} \\
  
 &  307 $\vert$ ~251 & 0.89 $\vert$ ~1.00 & \textbf{G} \\
 &  321 $\vert$ ~286 & 0.47 $\vert$ ~0.88 & \textbf{I} \\
 &  324 $\vert$ ~293 & 0.43 $\vert$ ~0.87& \textbf{Both} \\
 \bottomrule
\end{tabular}
}
\end{minipage}
\vspace{-6pt}
\end{table}

As an example, we show LfD for the \texttt{Franka-LiftCube} task. For each of the four settings of initial and desired object positions (fixed or random start and desired positions), we collect 2000 trajectories. Using these demonstrations, we train policies using Behavior Cloning (BC) and BC with an RNN policy (BC-RNN), and show their performance in~\tabref{tab:imitation}.

\subsection{Motion planning}
Motion planning is one of the well-studied domains in robotics. The traditional Sense-Model-Plan-Act (SMPA) methodology decomposes the complex problem of reasoning and control into possible sub-components.  
\simName supports doing this both procedurally and interactively via the GUI.

\paragraph{Hand-crafted policies} We create a state machine for a given task to perform sequential planning as a separate node in the \texttt{agent}. It provides the goal states for reaching a target object, closing the gripper, interacting with the object, and maneuvering to the next target position. We demonstrate this paradigm for several tasks in~\figref{fig:cloth-manip}. These hand-crafted policies can also be utilized for collecting expert demonstrations for challenging tasks such as cloth manipulation.

\begin{figure}[!t]
    \centering
    \minipage{0.45\linewidth}
        \includegraphics[width=0.95\linewidth]{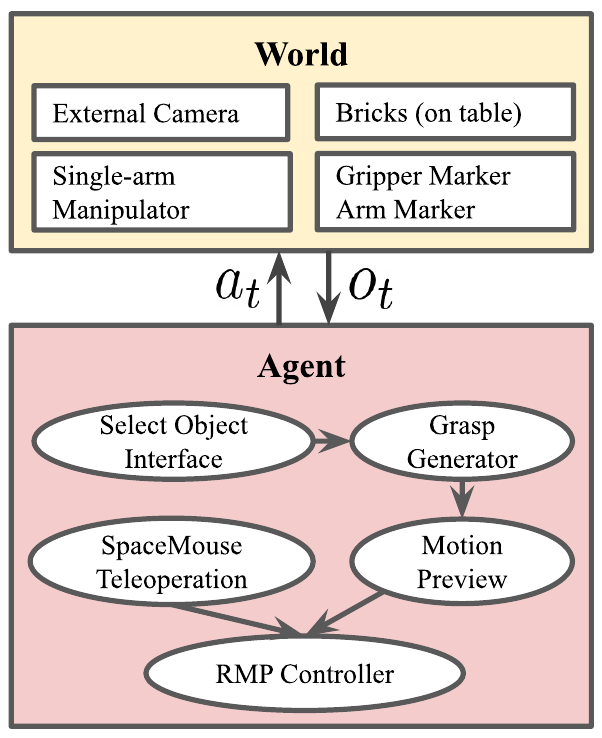}
    \endminipage
    \hfill
    \minipage{0.54\linewidth}
        \begin{tikzonimage}[width=0.48\linewidth]{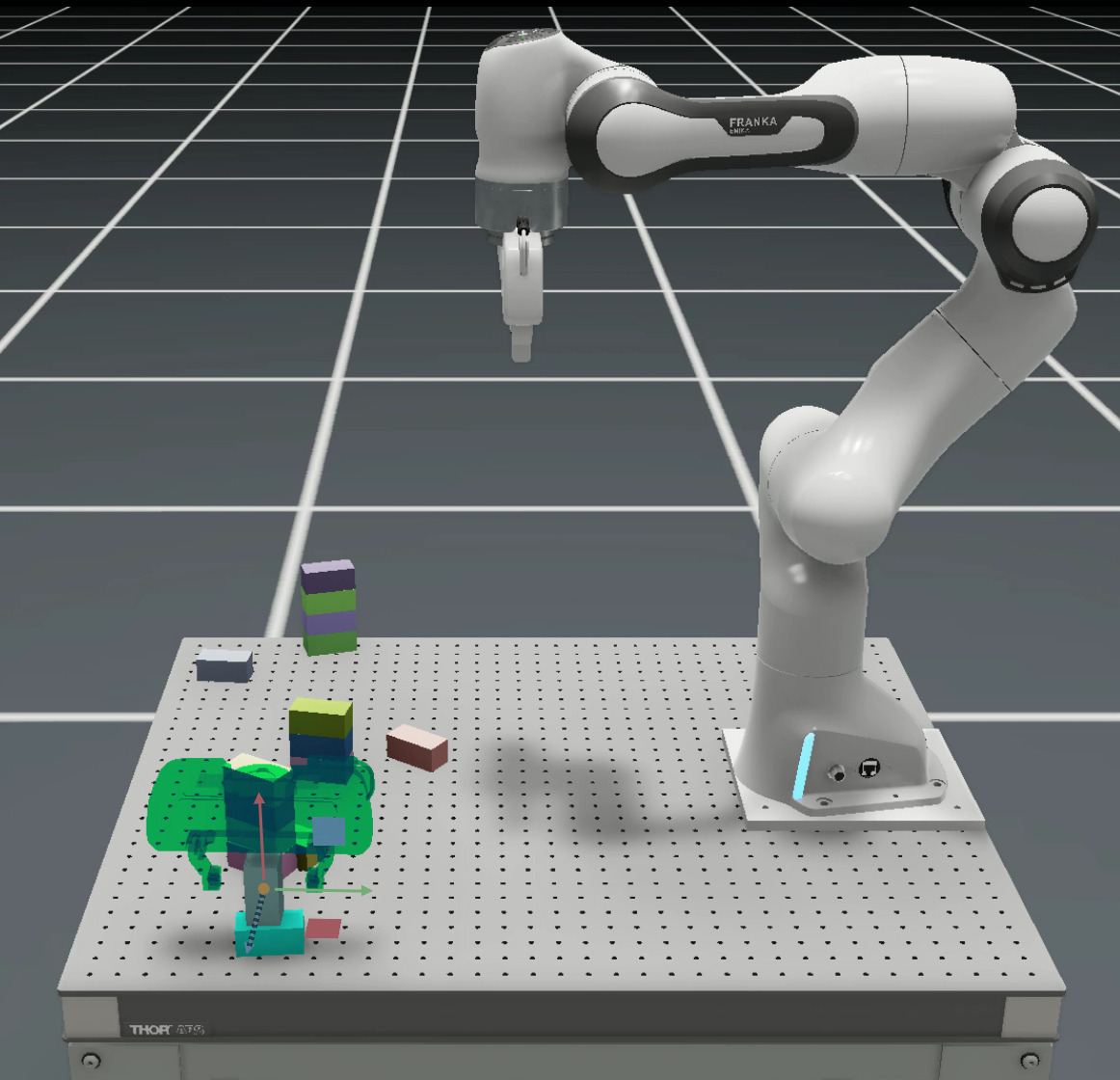}[image label]
        \node{1};
        \end{tikzonimage}
        \begin{tikzonimage}[width=0.48\linewidth]{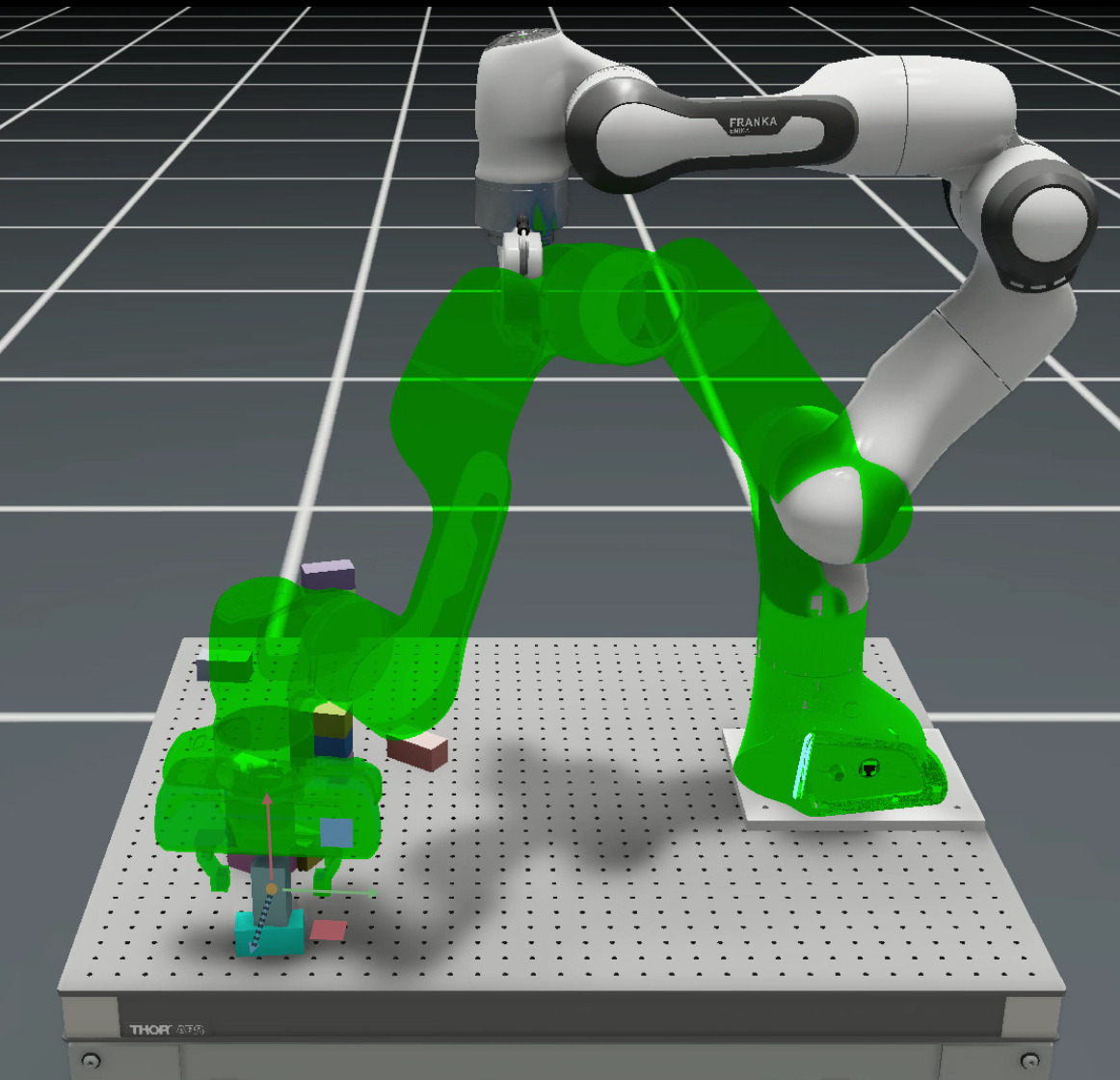}[image label]
        \node{2};
        \end{tikzonimage}
        
        \begin{tikzonimage}[width=0.48\linewidth]{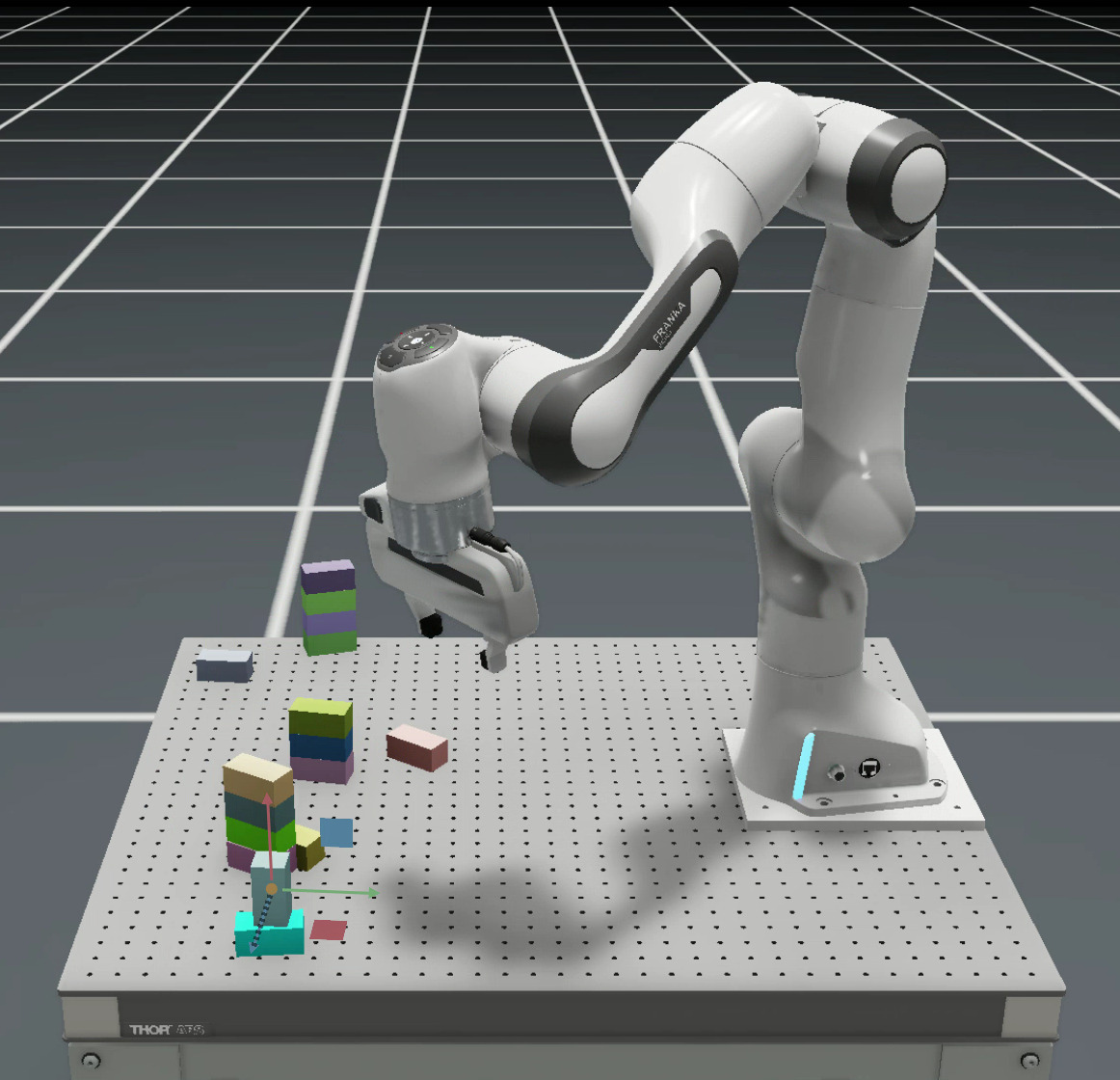}[image label]
        \node{3};
        \end{tikzonimage}
        \begin{tikzonimage}[width=0.48\linewidth]{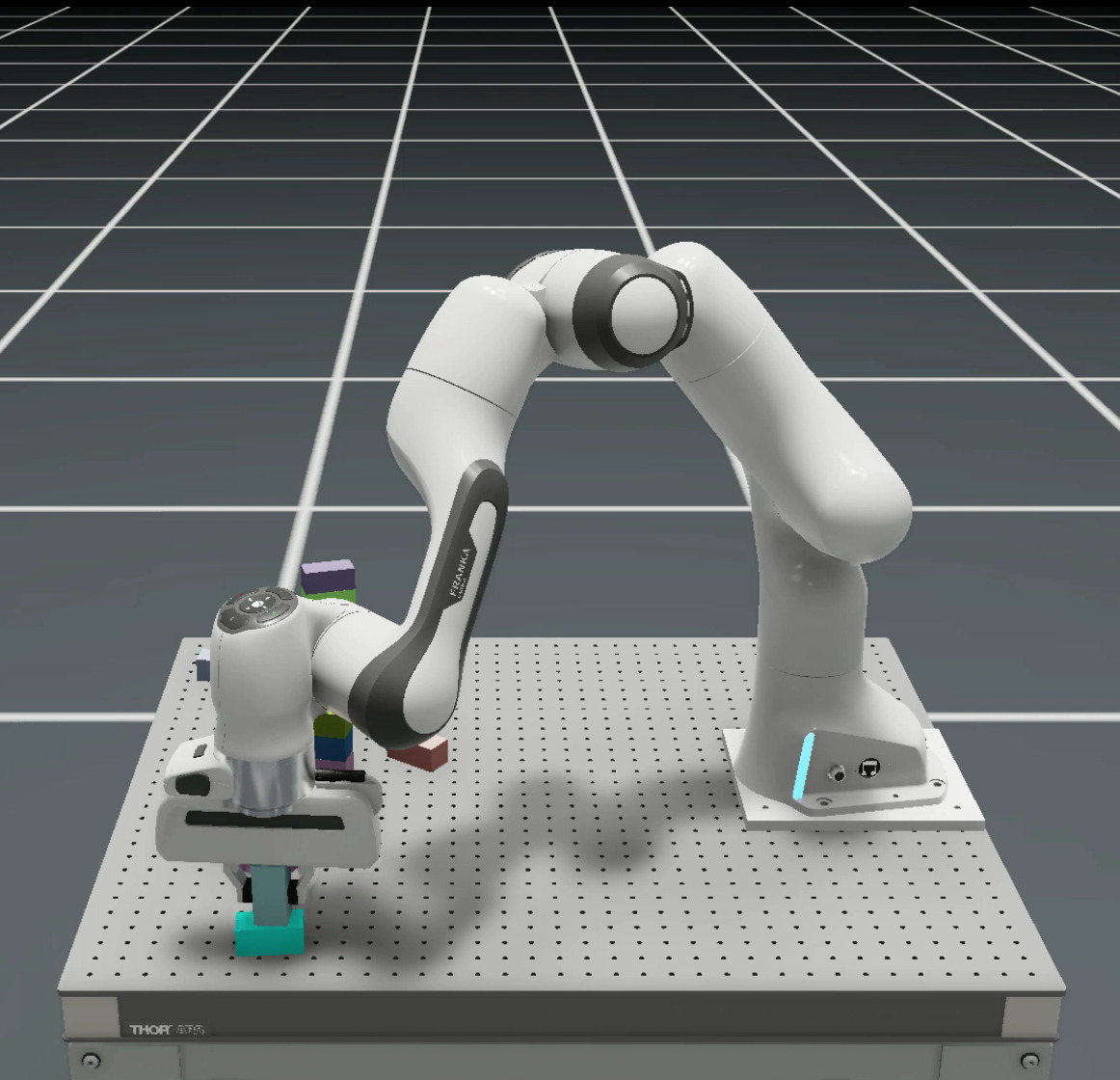}[image label]
        \node{4};
        \end{tikzonimage}
    \endminipage
    \caption{Interactive grasp and motion planning demonstration using \simName. The \texttt{World} comprises objects for table-top manipulation. The user can select an object from the GUI to grasp. This triggers an image-based grasp generator and allows previewing of the generated grasps and the robot motion sequence. The user can then choose the grasp and execute the motion on the robot.}
    \label{fig:franka_grasp_demonstration}
    \vspace{-9pt}
\end{figure}

\paragraph{Interactive motion planning} We define a system of nodes for grasp generation, teleoperation, task-space control, and motion previewing (shown in \figref{fig:franka_grasp_demonstration}). Through the GUI, the user can select an object to grasp and view the possible grasp poses and the robot motion sequences generated using the RMP controller. After confirming the grasp pose, the robot executes the motion and lifts the object. Following this, the user obtains teleoperation control of the robot. %

\begin{figure}[!t]
    \centering
    \minipage{0.63\linewidth}
        \centering
        \includegraphics[width=\textwidth]{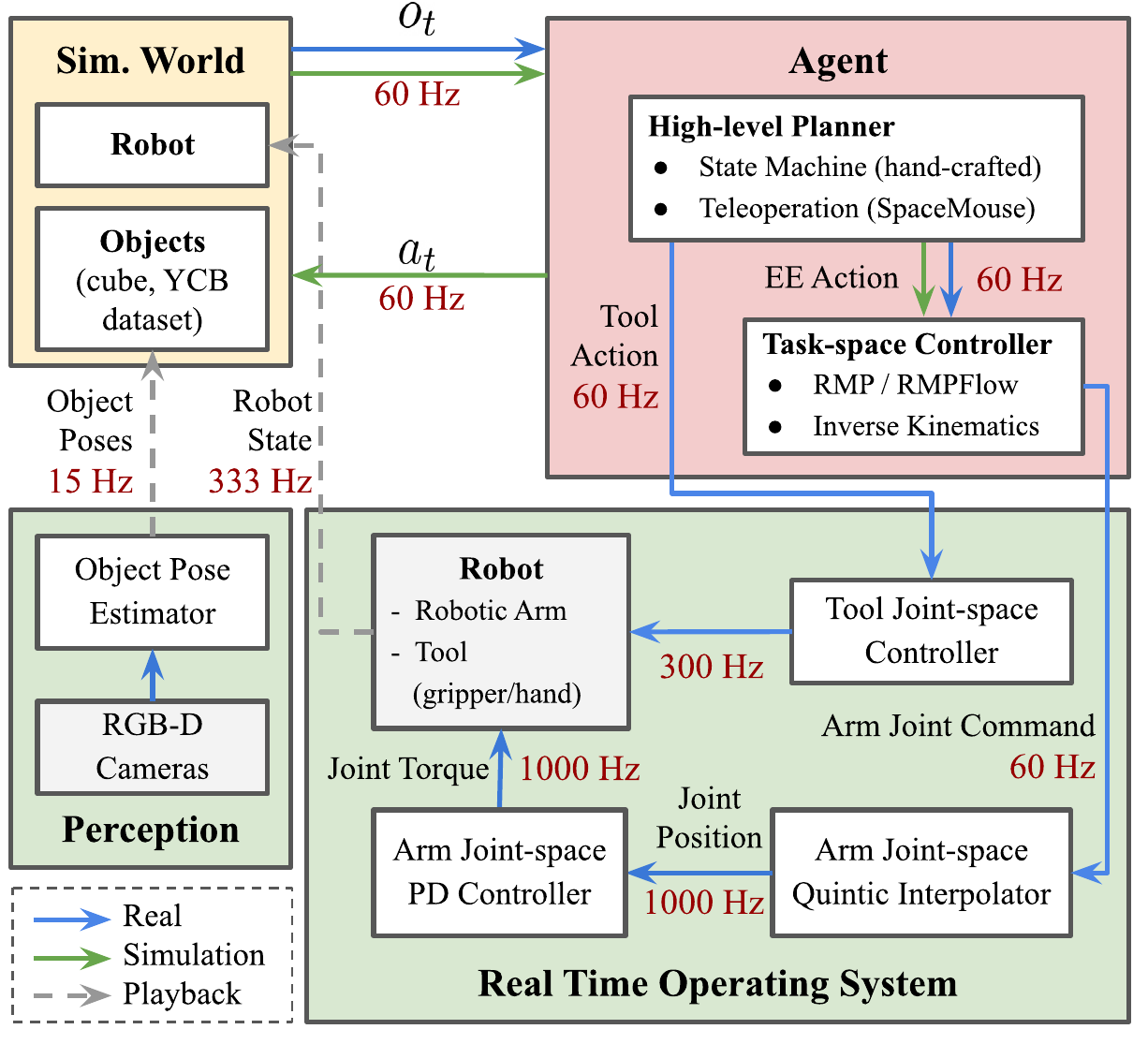}
    \endminipage
    \minipage{0.32\linewidth}
        \centering
        \includegraphics[width=0.94\textwidth]{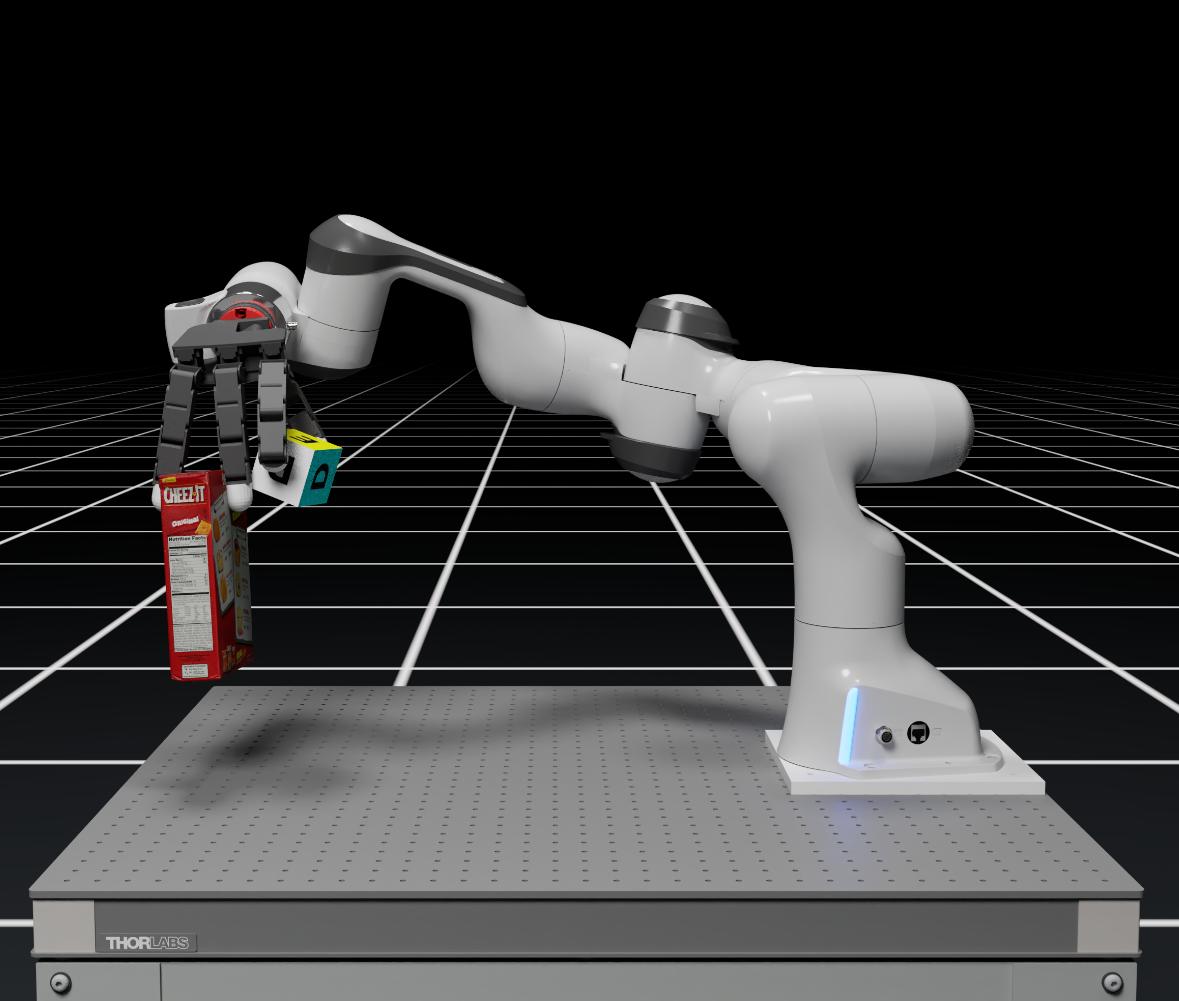}
        \includegraphics[width=0.94\textwidth]{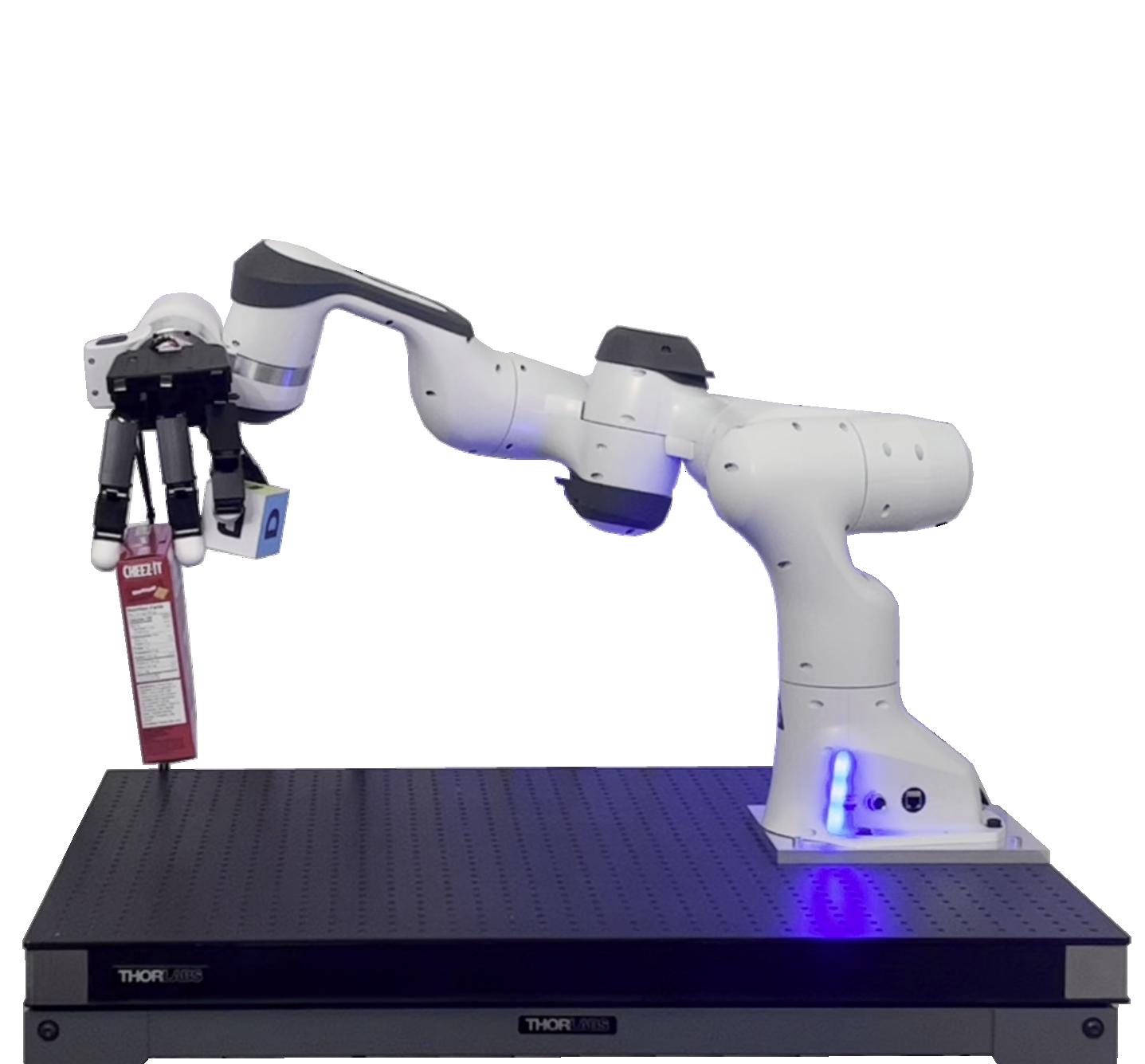}
    \endminipage
    \caption{Using the simulator as a digital twin to compute and apply the same commands on the simulated and real robot via ZMQ connection. We show the Franka Panda arm with an Allegro hand lifting two objects simultaneously, showcasing the realism in contact simulation. For videos, please check the \href{https://isaac-orbit.github.io/}{website}.
    }
    \label{fig:sim-real-allegro-lift}
    \vspace{-14pt}
\end{figure}

\subsection{Deployment on real robot}
\label{sec:real-robot-exp}

Deploying an \texttt{agent} on a real robot faces various challenges, such as dealing with real-time control and safety constraints. Different data transport layers, such as ROSTCP~\cite{quigley2009ros} or ZeroMQ (ZMQ)~\cite{hintjens2013zeromq}, exist for connecting a robotic stack to a real platform. We showcase how these mechanisms can be used with \simName to run policies on a real robot.

\paragraph{Using ZMQ} To maintain lightweight and efficient communication, we use ZMQ to send joint commands from \simName to a computer running the real-time kernel for Franka Emika robot. To abide by the real-time safety constraints, we use a quintic interpolator to upsample the 60 Hz joint commands from the simulator to 1000 Hz for execution on the robot (shown in~\figref{fig:sim-real-allegro-lift}).

We run experiments on two configurations of the Franka robot: one with the Franka Emika hand and the other with an Allegro hand. For each configuration, we showcase three tasks: 1) teleoperation using a Spacemouse device, 2) deployment of a state machine and 3) waypoint tracking with obstacle avoidance. The modular nature of the \texttt{agent} makes it easy to switch between different control architectures for each task while using the same interface for the real robot. %

\paragraph{Using ROS} A variety of existing robots come with their ROS software stack. In this demonstration, we focus on how policies trained using \simName can be exported and deployed on a robotic platform, particularly for the quadrupedal robot from ANYbotics, ANYmal-D.

\begin{figure}[!t]
    \centering
    \begin{tikzonimage}[width=0.325\linewidth,trim={0 0 400 0}, clip]{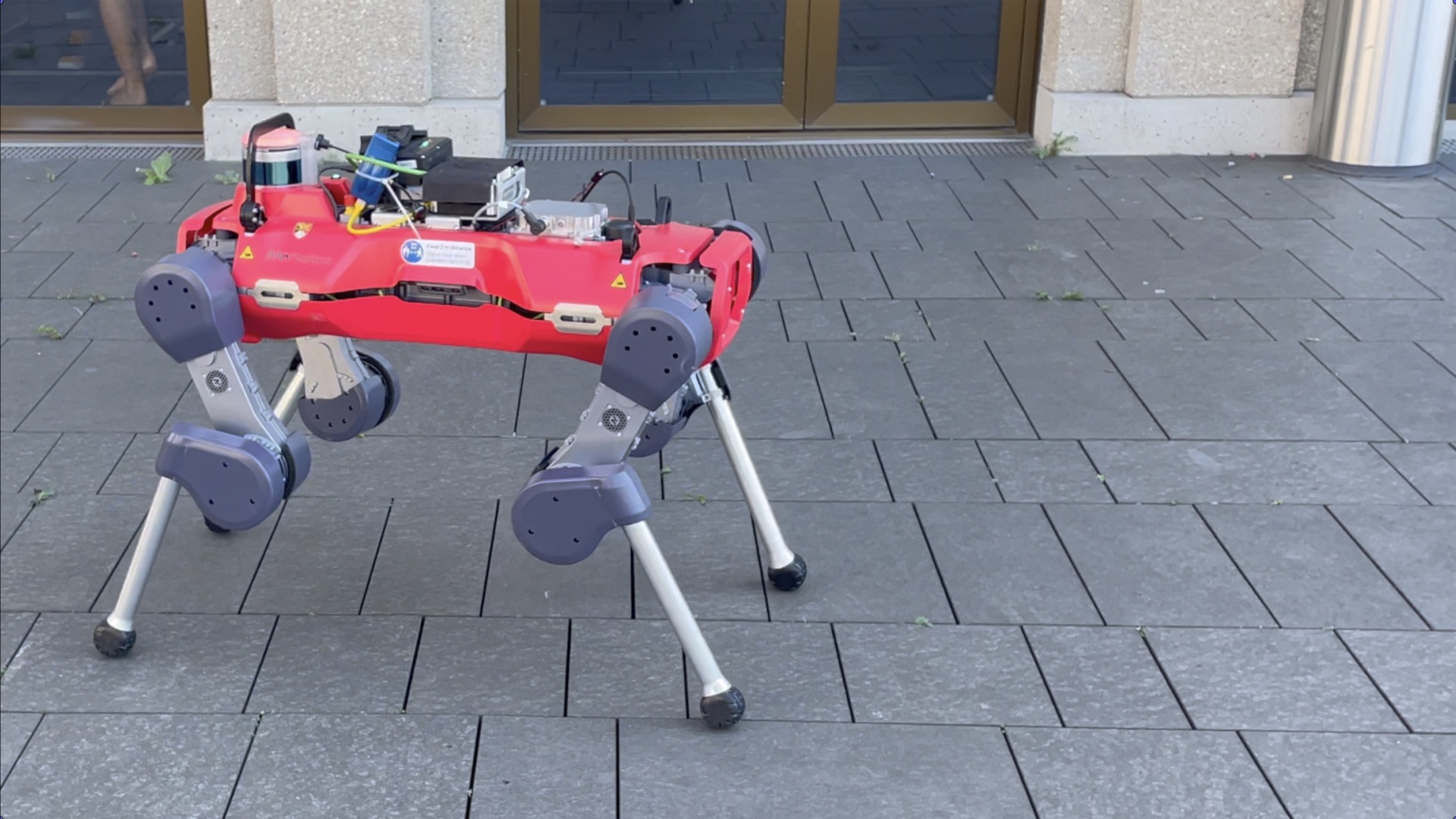}[image label]
        \node{1};
    \end{tikzonimage}
    \begin{tikzonimage}[width=0.325\linewidth,trim={0 0 400 0}, clip]{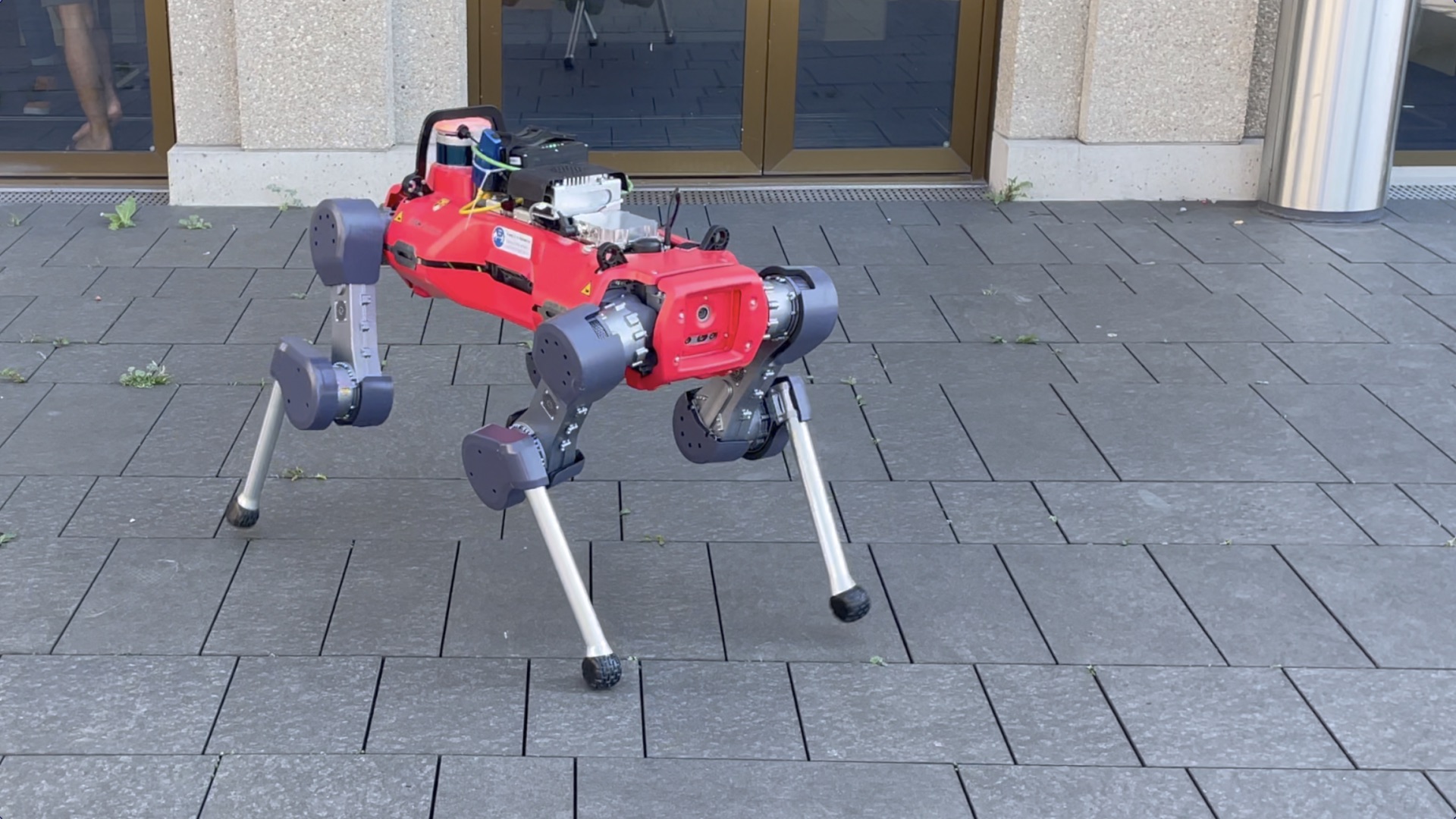}[image label]
        \node{2};
    \end{tikzonimage}
    \begin{tikzonimage}[width=0.325\linewidth,trim={0 0 400 0}, clip]{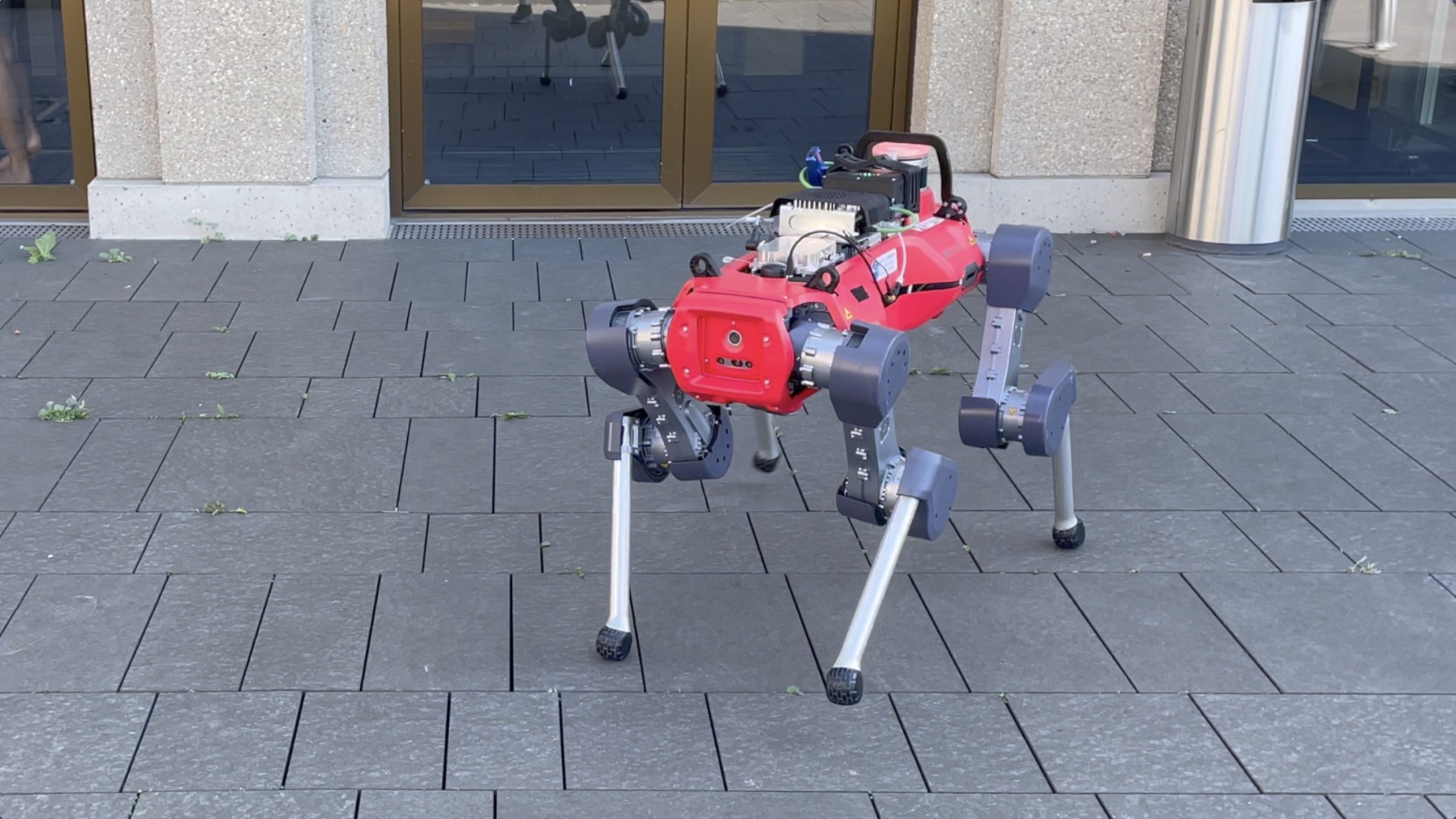}[image label]
        \node{3};
    \end{tikzonimage}
    \caption{Deployment of an RL policy on ANYmal-D robot using ROS connection (\href{https://youtu.be/zqpNogUAQBM}{video}). The policy is trained in simulation and runs at 50 Hz while the actuator net functions at 200 Hz.}
    \label{fig:real-anymal}
    \vspace{-12pt}
\end{figure}

We train a locomotion policy entirely in simulation using an actuator network~\cite{hwango2019actuator} for the legged base. To make the policy robust, we randomize the base mass ($22 \pm 5$ kg) and add simulated random pushes. We use the contact reporter to obtain the contact forces and use them in reward design. The learned policy is deployed on the robot using the ANYmal ROS stack, (\figref{fig:real-anymal}). This sim-to-real transfer indicates the viability of the simulated contact dynamics and its suitability for contact-rich tasks in \simName.

\begin{figure}
    \centering
    \includegraphics[width=\linewidth]{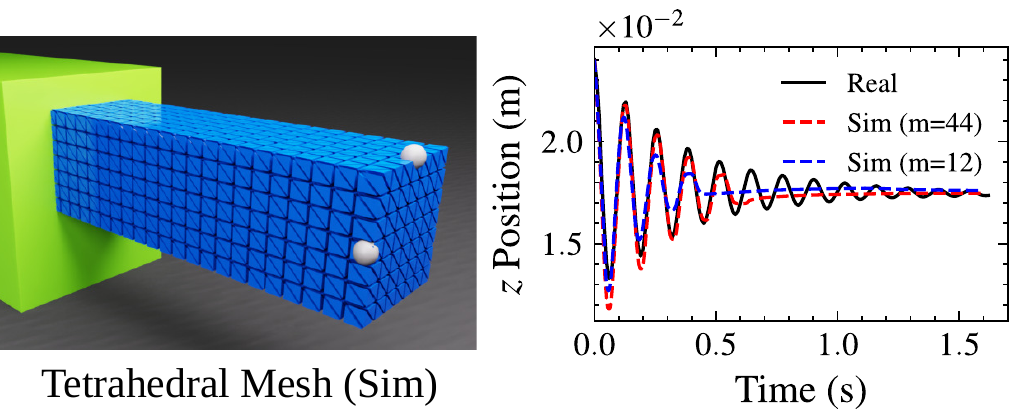}
    \caption{\rebuttal{Comparison of real and simulated data for a clamped beam scenario. The real data is collected using motion markers (shown on left). The simulated data is obtained using the FEM-solver~\cite{neohooken2021miles} in Isaac Sim over different hexahedral mesh resolutions (\textit{m}).}}
    \label{fig:beam-fem}
    \vspace{-15pt}
\end{figure}

\subsection{Simulation Evaluation and Comparison}
\label{sec:exp-sim-comp}

\rebuttal{
\paragraph{Evaluation of simulation accuracy}
Quantitatively measuring the accuracy of physics solvers is an arduous task due to the large variations and unknown mechanics of the real world. Thus, prior frameworks resort to qualitatively comparing the simulation to the physical world by rolling out the same action sequences~\cite{Lin2020softgym,gu2023maniskill2}. While \secref{sec:real-robot-exp} follows a similar practice to show realism in the rigid body simulation, we discuss the accuracy of deformable body simulation through a controlled experiment.
}

\rebuttal{
We consider a clamped beam made of highly deformable silicone elastomer. Following the setup used in~\cite{dubied2022sim}, we attach motion capture markers to the beam to collect deformation data
under gravity's effect. In the simulation, we create a similar scenario by attaching a soft beam to a rigid wall and setting the same material properties (Young's modulus, density, and incompressibility) as the physical beam. As shown in~\figref{fig:beam-fem}, we observe that the damped oscillations in the simulated data follow closely to the collected real-world data. This showcases the solver's accuracy~\cite{neohooken2021miles} and indicates its potential for sim-to-real deformable body manipulation.
}

\rebuttal{\paragraph{Comparison of simulation throughput} We compare the throughput of the tasks in \simName with those in other frameworks for rigid (robosuite, Maniskill2, IsaacGymEnvs) and deformable body interactions (DEDO). To ensure a fair comparison, we adapt the environments to have the same action space, simulation frequency, and control decimation. 
The evaluation is performed on a workstation with a 16-core AMD Ryzen 5950X, 64 GB RAM, and NVIDIA 3090RTX. 
}

\begin{figure}
    \centering
    \includegraphics[width=0.49\linewidth]{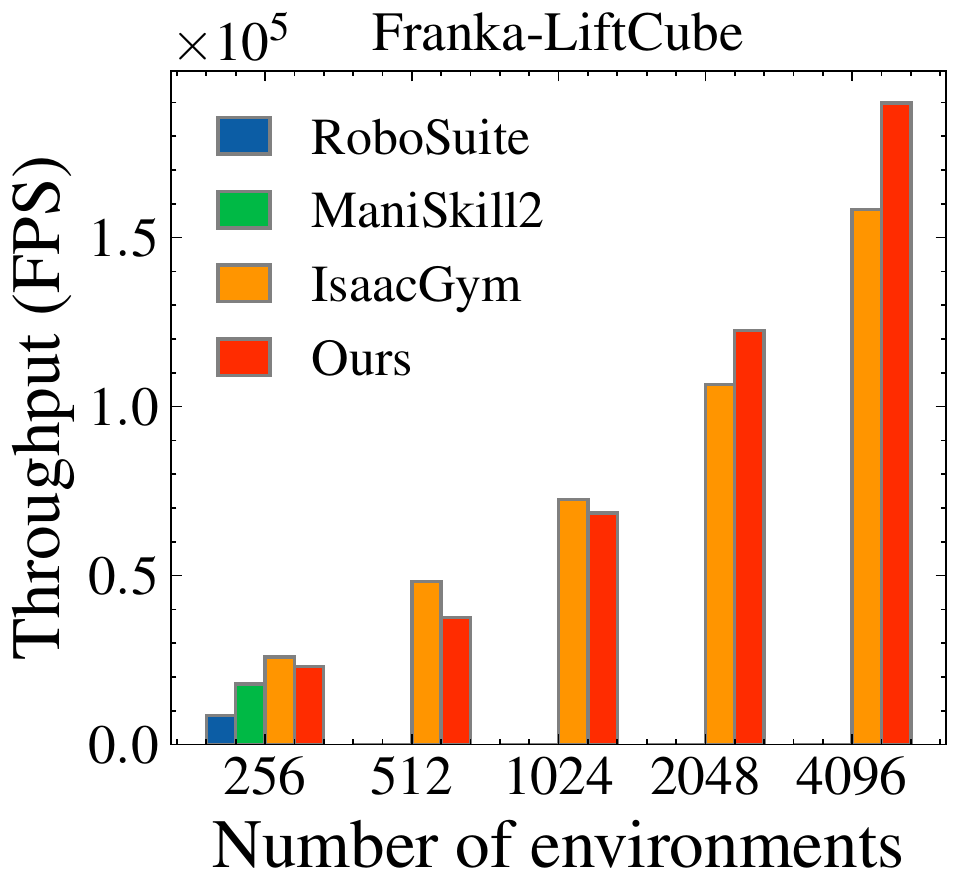}
    \includegraphics[width=0.43\linewidth]{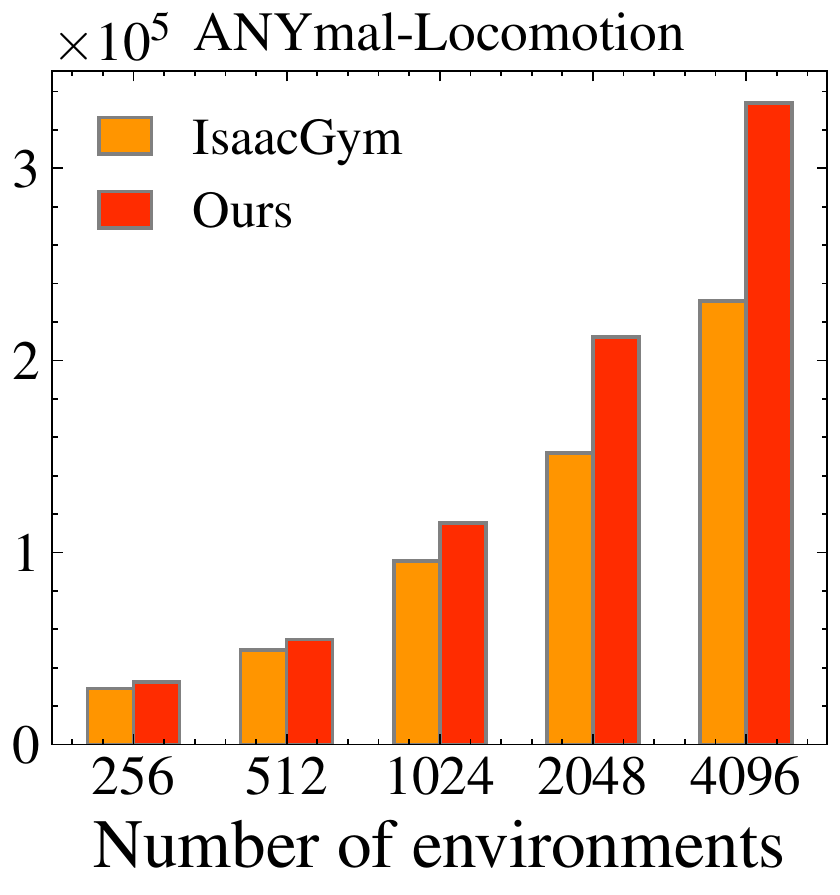}
    \caption{\rebuttal{Simulation throughput for rigid body tasks\protect\footnotemark. CPU-based vectorization is limited to the available memory and scales poorly compared to GPU-accelerated simulation. \simName performs at par with IsaacGym since they use the same physics engine.}}
    \label{fig:rigid-body-throughput}
    \vspace{-5pt}
\end{figure}

\begin{figure}
    \centering
    \includegraphics[width=0.523\linewidth]{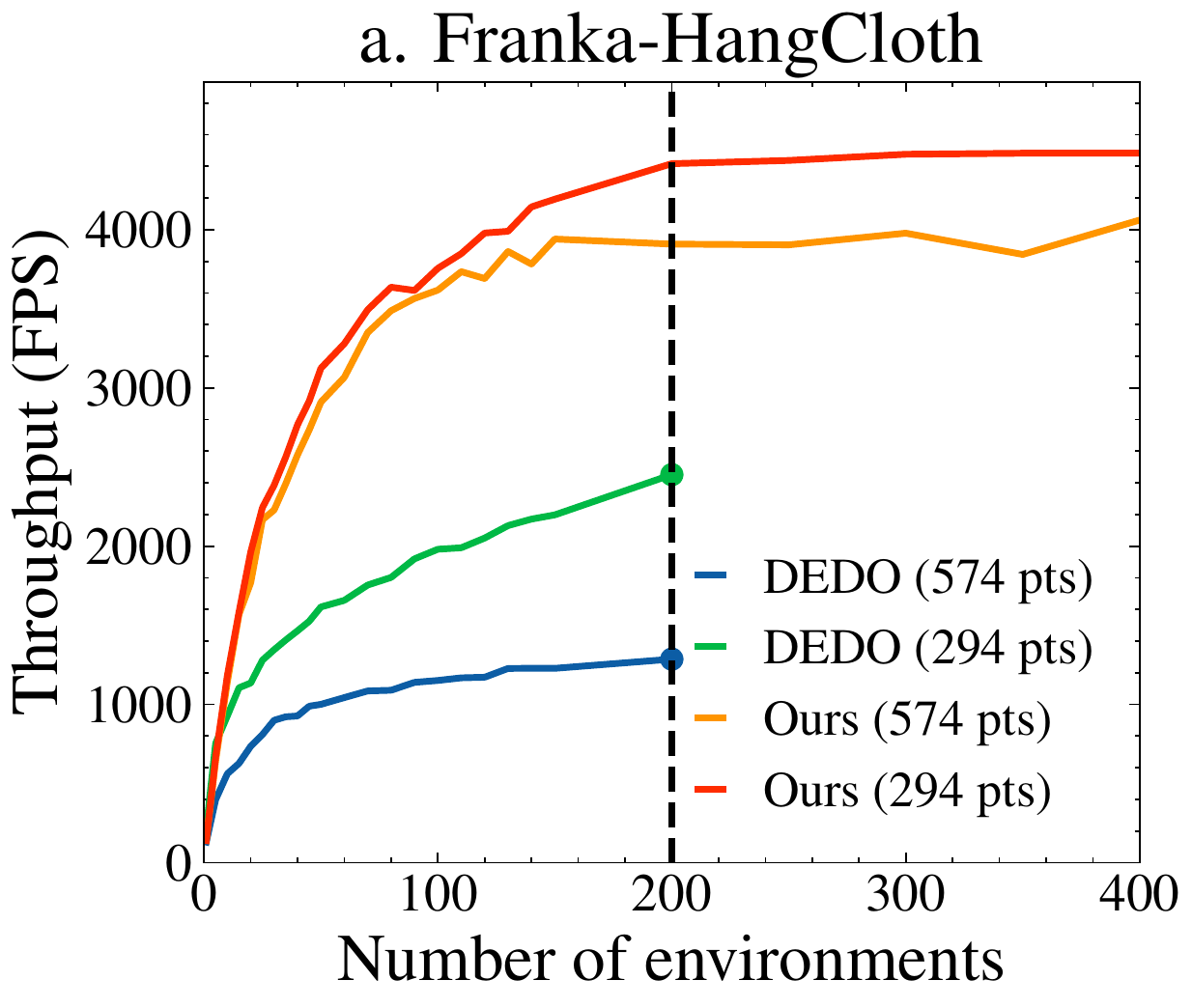}
    \includegraphics[width=0.457\linewidth,trim={23 0 0 0},clip]{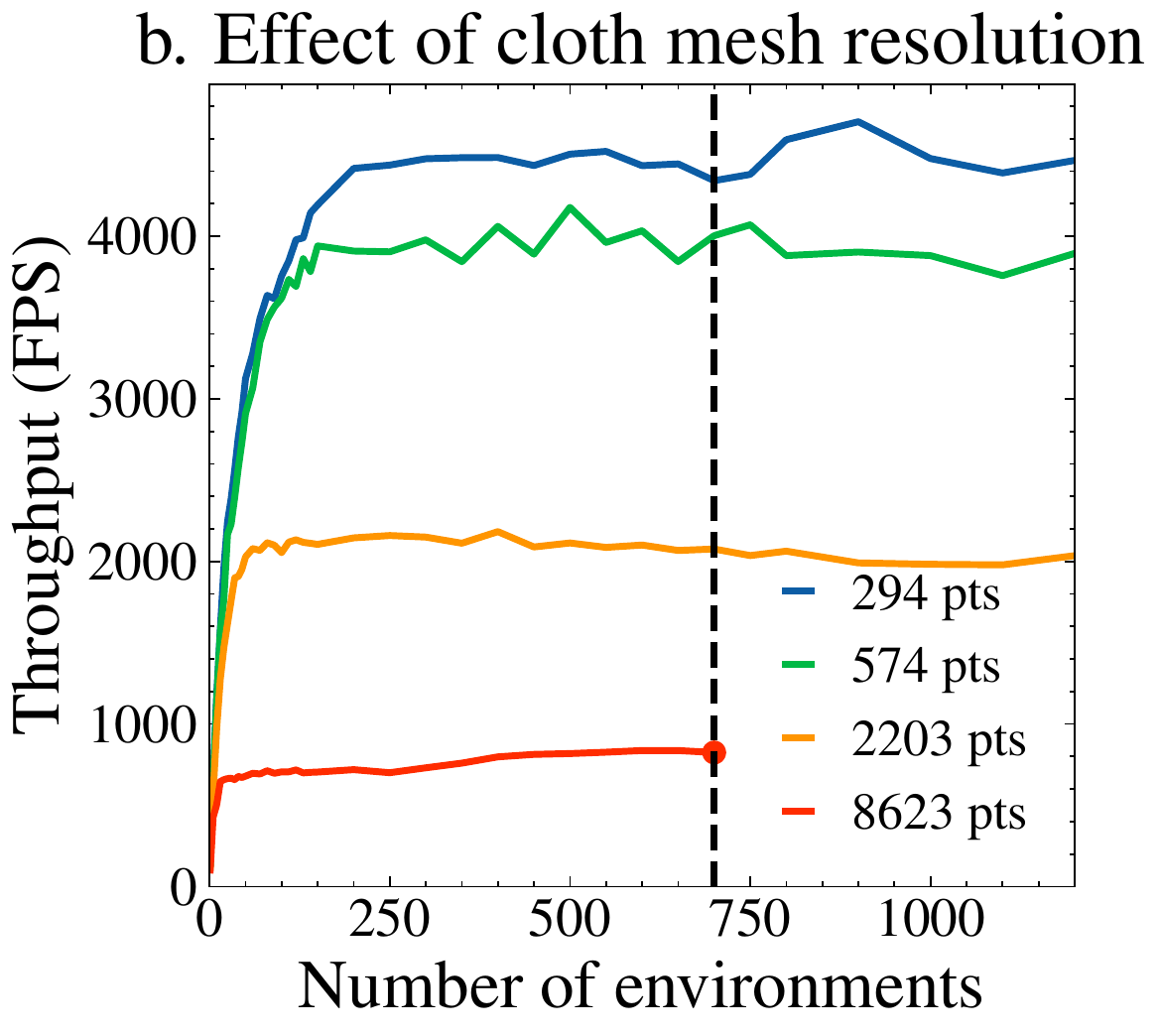}
    \caption{\rebuttal{Simulation throughput for cloth hanging task\textsuperscript{2}}. \simName obtains $3\times$ the throughput than DEDO while using PBD solver~\cite{pbf2013miles}. However, increasing the number of nodes or points (\textit{pts}) in the cloth mesh adversely affects its performance. The dotted line shows where the system ran out of memory.}
    \label{fig:cloth-throughput}
    \vspace{-15pt}
\end{figure}

\footnotetext{For rigid object environments, the numbers were computed using robosuite v1.4, ManiSkill2 v0.4, Isaac Gym Preview 4, and Isaac Sim 2022.1. For the cloth environment, we used DEDO v0.1 and Isaac Sim 2022.2.}

\rebuttal{
Frameworks that rely on CPU-vectorization (\cite{zhu2020robosuite,antonova2021dedo,gu2023maniskill2}) show an increase in throughput with the number of environments. However, at around 200-300 environments, their programs crash due to insufficient memory. In contrast, GPU-based parallelization scales better to a larger number of environments and achieves a throughput of ${\sim}10$x faster for rigid body environments (\figref{fig:rigid-body-throughput}) and ${\sim}3$x faster for deformable body environments (\figref{fig:cloth-throughput}). For the cloth task, we also examine the impact of using meshes of different resolutions on the throughput. We observe that a higher mesh resolution produces more accurate simulation but requires more computation time. To address this challenge, we intend to provide best practices for tuning the simulation and various sim-ready assets.
}

%% file: sections/6-conclusion.tex
\section{Discussion}

In this paper, we proposed \simName: an interactive and intuitive framework to simplify environment designing, enable easy task specifications, and lower the entry barrier into robotics and robot learning. \simName exploits the latest state-of-the-art simulation capabilities through Isaac Sim and extends them further to incorporate different actuator and sensor noise models into the simulation, and advance sensors, actuators, and motion generators at varying operating frequencies. It readily comes with different robotic platforms, sensors, CPU and GPU-based motion generators, and benchmark tasks that aim to provide a batteries-included experience for roboticists. The breadth of environments and robotic paradigms possible, as demonstrated in part in~\secref{sec:features} and~\secref{sec:workflows}, make \simName useful for a broad set of research questions in robotics. Through experiments, we show a significant throughput improvement on tasks designed in \simName with respect to those in other frameworks, and its potential to facilitate sim-to-real transfer.

By open-sourcing this framework\footnote{
\nvidia Isaac Sim is free with an \href{https://www.nvidia.com/en-us/omniverse/download/}{individual license}. \simName is open-sourced on GitHub and available at~\href{https://isaac-orbit.github.io/}{https://isaac-orbit.github.io}.%
}, 
we aim to reduce the overhead for developing new applications and provide a unified platform for robot learning research. As we continue to enhance and incorporate more features into the framework, we encourage researchers to contribute to transforming it into a comprehensive solution for robotics research.

\section{Future Work}

\simName can notably simulate physics at up to 125,000 FPS; however, camera rendering is currently bottlenecked to a total of 270 FPS for ten cameras rendering $640 \times 480$ images on an RTX 3090. While this number is comparable to other frameworks, we are actively improving the rendering speed through GPU-based acceleration.

While our experiments demonstrate the effectiveness of rigid-contact modeling and FEM for soft bodies, \rebuttal{quantitatively studying the fidelity of the entire simulator (such as rendering, sensors, and physics) remains an area for future exploration. It is important to note that robotics research, particularly in deformable-body manipulation, has infrequently used sim-to-real due to difficulties in achieving fast and accurate simulation and realistic rendering. We believe that \simName can help address these challenges and facilitate answering open research questions in these fields.} 

\rebuttal{Further enhancements to the framework include integrating tactile sensors and 6-axis force-torque sensors. Additionally, we plan to add support for loading assets directly in their native formats (such as URDF and OBJ) instead of USDs to make the framework more versatile and user-friendly.}

%% file: sections/7-appendix.tex
\section*{Appendix A: Author Contributions}

The following lists author contributions by the type:

\begin{itemize}
    \item \textit{Designed and built the core infrastructure:} M. Mittal, D. Hoeller, N. Rudin
    \item \textit{Integrated robots into the simulator:} M. Mittal, J. Liu
    \item \textit{Implemented motion generators:} M. Mittal, J. Liu
    \item \textit{Added environments for locomotion:} M. Mittal, D. Hoeller, N. Rudin
    \item \textit{Added environments for rigid object manipulation:} M. Mittal, J. Liu, A. Yuan
    \item \textit{Added environments for deformable object manipulation:} C. Yu, Q. Yu
    \item \textit{Ran workflows evaluations:} J. Liu, C. Yu, Q. Yu
    \item \textit{Performed throughput comparisons:}  C. Yu, M. Mittal
    \item \textit{Conducted sim-to-real experiments on Franka arm:} R. Singh, J. Liu
    \item \textit{Conducted sim-to-real experiments on ANYmal robot:} M. Mittal, D. Hoeller
    \item \textit{Provided NVIDIA Isaac Sim support:} Y. Guo, H. Mazhar, B. Babich, G. State
    \item \textit{Managed or advised on the project:} A. Mandlekar, M. Hutter, A. Garg
    \item \textit{Wrote the paper:} M. Mittal, A. Garg
\end{itemize}

\section*{Appendix B: Motivation behind \simName}

Over the years, NVIDIA has developed a number of tools for robotics and AI. These tools leverage the power of GPUs to accelerate the simulation both in terms of speed and realism. They show great promise in the field of simulation technology and are being used by many researchers and companies worldwide.

\textbf{Isaac Gym}~\cite{makoviychuk2021isaac} provides a high-performance GPU-based physics simulation for robot learning. It is built on top of PhysX which supports GPU-accelerated simulation of rigid bodies and a Python API to directly access physics simulation data. Through an end-to-end GPU pipeline, it is possible to achieve high frame rates compared to CPU-based physics engines. The tool has been used successfully in a number of research projects, including legged locomotion~\cite{rudin2022learning}, in-hand manipulation~\cite{allshire2021transferring}, and industrial assembly~\cite{narang2022factory}.

Despite Isaac Gym's success and adoption, it is not designed to be a general-purpose simulator for robotics. For instace, it does not include interaction between deformable and rigid objects, high-fidelity rendering, and support for ROS. The tool has been primarily designed as a preview release to showcase the capabilities of the underlying physics engine. With the release of Isaac Sim, NVIDIA is building a general-purpose simulator for robotics and has integrated the functionalities of Isaac Gym into Isaac Sim.

\textbf{Isaac Sim} is a robot simulation toolkit built on top of Omniverse, which is a general-purpose platform that aims to unite complex 3D workflows. Isaac Sim leverages the latest advances in graphics and physics simulation to provide a high-fidelity simulation environment for robotics. It supports ROS/ROS2, various sensor simulations, tools for domain randomization, and synthetic data creation. Overall, it is a powerful tool for roboticists and is a huge step forward in the field of robotics simulation.

With the release of the above two tools, NVIDIA also released an open-sourced set of RL environments called \href{https://github.com/NVIDIA-Omniverse/IsaacGymEnvs}{IsaacGymEnvs} and \href{https://github.com/NVIDIA-Omniverse/OmniIsaacGymEnvs}{OmniIsaacGymEnvs}, which use Isaac Gym and Isaac Sim respectively. These environments have been designed to display the capabilities of the underlying simulators and provide a great starting point to understand what is possible with the simulators for robot learning. However, these repositories are constrained by their limited integration with different RL libraries (beyond RL-Games), lack of modularity for scalable environment design, and inadequate support for paradigms beyond RL. It is these deficiencies that sparked the development of the \simName framework, aiming to bridge this gap and empower researchers and developers alike.

% \section*{Appendix C: NVIDIA Isaac Ecosystem}